\documentclass[journal]{IEEEtran}
\usepackage{cite}

\usepackage{color}

\usepackage{amsmath}

\usepackage{algorithm}
\usepackage{algorithmic}
\usepackage{bm}
\usepackage{latexsym}
\usepackage{amsthm}
\usepackage{url}
\usepackage{amsfonts}
\usepackage{amssymb}
\usepackage{indentfirst}
\usepackage{float}
\ifCLASSINFOpdf
   \usepackage[pdftex]{graphicx}
   \graphicspath{{../pdf/}{../jpeg/}}
   \DeclareGraphicsExtensions{.pdf,.jpeg,.png}
\else
   \usepackage[dvips]{graphicx}
   \graphicspath{{../eps/}}
   \DeclareGraphicsExtensions{.eps}
\fi

\usepackage{array}
\usepackage{multirow}

\ifCLASSOPTIONcompsoc
  \usepackage[caption=false,font=normalsize,labelfont=sf,textfont=sf]{subfig}
\else
  \usepackage[caption=false,font=footnotesize]{subfig}
\fi

\begin{document}

\title{Asymmetric CNN for image super-resolution}

\author{Chunwei ~Tian,
        Yong ~Xu$^*$,~\IEEEmembership{Senior Member,~IEEE,}
        Wangmeng ~Zuo,~\IEEEmembership{Senior Member,~IEEE,}
        Chia-Wen ~Lin$^*$,~\IEEEmembership{~Fellow,~IEEE}
        and David ~Zhang,~\IEEEmembership{~Life Fellow,~IEEE}
\thanks{This work was supported in part by the National Nature Science Foundation of China under Grant No. 61876051, in part by the Shenzhen Municipal Science and Technology Innovation Council under Grant JCYJ20180306172101694, and in part by Ministry of science and Technology, Taiwan, under Grant 110-2634-F-007-015-. (Corresponding author: Yong Xu (Email: yongxu@ymail.com) and Chia-Wen Lin (Email: cwlin@ee.nthu.edu.tw).)}
\thanks{Chunwei Tian is with the Bio-Computing Research Center, Harbin Institute of Technology, Shenzhen, Shenzhen, 518055, Guangdong, China. He is also with Shenzhen Key Laboratory of Visual Object Detection and Recognition, Shenzhen, 518055, Guangdong, China. (Email: chunweitian@163.com.)}
\thanks{Yong Xu and Wangmeng Zuo are with the School of Computer Science and Technology, Harbin Institute of Technology, Harbin, 150001, Heilongjiang, China. And they are also with the Peng Cheng Laboratory, Shenzhen, 518055, Guangdong, China. Additionally, Yong Xu is with Shenzhen Key Laboratory of Visual Object Detection and Recognition, Shenzhen, 518055, Guangdong, China.
(Email: yongxu@ymail.com and wmzuo@hit.edu.cn).}
\thanks{Chia-Wen Lin is with the Department of Electrical Engineering and the Institute of Communications Engineering, National Tsing Hua University, Hsinchu, Taiwan (Email: cwlin@ee.nthu.edu.tw)}
\thanks{David Zhang is with the School of Science and Engineering, The Chinese University of Hong Kong (Shenzhen), Shenzhen, 518172, Guangdong, China. And  he  is  also  with  the  Shenzhen  Institute  of  Artificial  Intelligence  and Robotics for Society, Shenzhen, China (Email: davidzhang@cuhk.edu.cn)}}
\maketitle


\begin{abstract}
Deep convolutional neural networks (CNNs) have been widely applied for low-level vision over the past five years. According to nature of different applications, designing appropriate CNN architectures is developed. However, customized architectures gather different features via treating all pixel points as equal to improve the performance of given application, which ignores the effects of local power pixel points and results in low training efficiency. In this paper, we propose an asymmetric CNN (ACNet) comprising an asymmetric block (AB), a memory enhancement block (MEB) and a high-frequency feature enhancement block (HFFEB) for image super-resolution. The AB utilizes one-dimensional asymmetric convolutions to intensify the square convolution kernels in horizontal and vertical directions for promoting the influences of local salient features for SISR. The MEB fuses all hierarchical low-frequency features from the AB via residual learning (RL) technique to resolve the long-term dependency problem and transforms obtained low-frequency features into high-frequency features. The HFFEB exploits low- and high-frequency features to obtain more robust super-resolution features and address excessive feature enhancement problem. Additionally, it also takes charge of reconstructing a high-resolution (HR) image. Extensive experiments show that our ACNet can effectively address single image super-resolution (SISR), blind SISR and blind SISR of blind noise problems. The code of the ACNet is shown at https://github.com/hellloxiaotian/ACNet.
\end{abstract}


\IEEEpeerreviewmaketitle

\section{Introduction}
\IEEEPARstart{S}{ingle} image super-resolution (SISR) is exploited to estimate a high-quality (also called high-resolution, HR) image via given degraded low-resolution (LR) image. It has been found in many applications, i.e., identity recognition \cite{liang2019novel} and medical diagnosis \cite{shi2013cardiac}. Generally speaking, the LR and HR images are characterized via a degradation model 
$y = x{ \downarrow _s}$, where  $y$ and $x$ are the LR and HR images, respectively. $\downarrow _s$ denotes downsampling operation with scale factor of $s$. According to the equation, we can see that the SISR task is an ill-posed problem. Thus, detailed information of the LR image is very important for SISR. To alleviate this problem, based on extra information techniques were proposed \cite{du2019improved}. For instance, relative displacements in image sequences can use predict idea to enhance the pixels of the unclear image\cite{irani1991improving}. Besides, image prior-gradient profile prior can show the shape and sharpness of image gradients to obtain more edge information for SISR \cite{sun2008image}. Moreover, the simultaneous use of adaptive regularization and learning-based super-resolution can eliminate the effect of the noise and mine more high-frequency information in the compression scenario to address multi-degradation task
\cite{xiong2010robust}. Additionally, nearest-neighbor patch method enlarged the influences of local pixel points via relation of different areas from the given LR image to
achieve the high-resolution image \cite{chang2004super}. In terms of improving the efficiency, sparse representation technique was also a good choice in SISR \cite{gao2014novel,alvarez2018imageb}. For instance, the combination of discrete wavelet transform, principal components analysis and sparse representation can reduce the information dimension to obtain more better expression carrier of the goal task for SISR \cite{alvarez2018image}. Besides, there were also other popular super-resolution methods, i.e., dictionary learning \cite{yang2016single} and random forest \cite{schulter2015fast}. Although these super-resolution methods made tremendous effort to promote the performance, they still suffered from some limitations: previous researches tended to boost the super-resolution performance via complex optimization methods. That may result in low execution speed. Another side effect is that these methods referred to manual setting parameters to obtain better results of SISR. Thereby, the more important thing is that a tool with power self-learning ability is critical to recover the HR image. 

To overcome these drawbacks, CNNs with powerful expressive capability for SISR were proposed \cite{tian2020image}. A three-layer CNN for SISR as well as SRCNN was found \cite{dong2014learning}. As the pioneer, the SRCNN first up-sampled LR image as input of the network through a end-to-end architecture to obtain the HR image. Although it was simple and shallow, its depth limited super-resolution performance. After that, many good efforts tried to make a trade-off between good performance and network depth. For instance, the residual connections had a good effect on resolving these issues. A very deep SR network (also regarded as VDSR) \cite{kim2016accurate} started from stacking multiple layers to increase the depth and used the skip connection to extract more robust SR features. Alternatively, a recursive operation had similar function to promote the quality of the unclear image. Kim et al. \cite{kim2016deeply} gathered hierarchical features by recursive learning technique without referring to additional parameters for preventing overfitting and addressing resource-constrained problem. Tai et al. \cite{tai2017image} extended this recursive mechanism through combining global and local information to strengthen the expressive ability of deep networks in low-level version applications. 
In terms of reducing the difficulty of training and handling gradient vanishing phenomenon, a symmetric skip fused into an encoder-decoder network was presented to remove the noise and recover the image details \cite{mao2016image}. However, these methods depended on bicubic interpolation operation to amplify given low-resolution image the same as the HR image, which brought great budget for training a super-resolution model. Also, most of these networks merged different obtained features via treating all pixels as equal to promote super-resolution performance, which may increase the effects of non-critical feature points and training cost.

In this paper, we propose an asymmetric CNN (also treated as ACNet) containing an AB, a MEB and a HFFEB for SISR. The AB enhances the effects of local key points on SISR via one-dimensional asymmetric convolutions in horizontal and vertical directions rather than treating all pixel points as equal.
To solve long-term dependency issue of deep network, the MSB merges all hierarchical low-frequency features from the AB through the RL technique to boost the memory ability of shallow layers on deep layers. Moreover, the MSB can use a flexible up-sampling mechanism to obtain high-frequency features for SISR and blind SISR. After that, taking into account sudden shock from up-sampling mechanism, the HFFEB 
fuses low- and high-frequency information to obtain more robust super-resolution (SR) features for restoring the high-quality image. The extended experiments describe that our ACNet performs well against state-of-the-art techniques, such as a lightweight enhanced super-resolution CNN (LESRCNN) \cite{tian2020lightweight} in terms of both quantitative and qualitative evaluations for SISR, blind SISR (i.e., a SR model for unknown scale factors) and multiple degradation (i.e., blind SISR of blind noise as well a SR model for unknown scales factor with unknown noise). 

The main contributions of our ACNet are shown as follows:

(1) We present a multi-level feature fusion mechanism by fusing hierarchical low-frequency features and high-frequency features to well resolve long-term dependency issue and prevent performance degradation from upsampling mechanism.   

(2) We propose an asymmetric architecture to enhance the effects of local key points for obtaining salient low-frequency features in SISR.  

(3) A flexible up-sampling mechanism  can make the proposed network resolve SISR, blind SISR and blind SISR of blind noise tasks.

The rest of this paper is illustrated as follows. Section 2 summaries the related work of deep CNNs for SISR, asymmetric convolutions and multi-level feature fusion on image super-resolution. Section 3 presents the proposed method. Section 4 illustrates the extended experiments, gives the principles and rationalities of the proposed key techniques and shows the experimental performance in SISR. Section 5 concludes the article.  
\section{Related work}
\subsection{Deep CNNs for SISR}
Due to powerful representation capability, CNNs have obtained great success in low-level vision tasks, especially SISR \cite{tian2019deep}. Dong et al. \cite{dong2014learning} utilized sparse coding mechanism to guide the CNN for obtaining the high-quality image. After that, numerous variants of CNNs were developed to improve the super-resolution performance, accelerate the training efficiency and handle complex low-level vision task. 

To promote super-resolution effect, researchers usually increase the depth or width of network to enlarge the receptive field of the network for mining more information. For instance, Fan et al. \cite{fan2018compressed} utilized a lot of multi-scale feature fusion mechanisms to increase the width of CNN for capturing more complete structure information. However, deeper and wider architectures may cause a larger amount of computation resource and higher memory consumption. To address this problem, Zhang et al. \cite{zhang2018dcsr} fused dilated convolutions into the CNN to the expand receptive field without referring to additional parameters and computational complexity. Alternatively, the discriminative learning method integrated into model-based optimization was a good tool to efficiently recover texture features of the high-definition image \cite{zhang2017learning}. Additionally, signal processing idea was beneficial to boost the pixels of the low-resolution image for obtaining the HR image \cite{liu2018multi}.

 In improving the training efficiency for a SR model, reducing the complexity of deep networks is common way, which includes two categories in general: decrease the number of training data and compress the network. For the first method, exploiting LR as input to train a SR model was a good choice. For instance, Dong et al. \cite{dong2016accelerating} used LR input and upscaling operations at the final layer of the network rather than up-sampling HR as the input for predicting the SR image. For the second method, dividing big filters into small filters had important effect on SISR. For instance, Ahn et al. \cite{ahn2018fast} extended convolution of $1 \times 1$ to the CNN to distill more useful information for improving execution speed of obtaining a HR image. Besides, splitting channels by group convolutions combined attention mechanism to facilitate more detailed information from structures, textures, and edges in SISR \cite{hui2019lightweight}. 
 
For dealing with complex low-level vision task, step-by-step mechanism was proposed. For instance, Zhang et al. \cite{zhang2018gated} presented a three-step method to tackle the LR image. The first step aimed to recover a clean LR image from the blurry LR image. The second step
used the LR input to extract high-dimensional features. The third step used attention mechanism \cite{tian2020attention}, obtained high-dimensional features and the clean LR image to enhance the high-frequency information and obtain clearer image. Additionally, it is known that external information played an important role in handling complex corrupted images, such as real noisy image \cite{du2018robust,wang2020weighted}, rainy image \cite{ren2019progressive}, foggy images \cite{ren2020single} and LR image \cite{yang2016single}. Inspired by that, Zhang et al. \cite{zhang2019deep} used half quadratic splitting algorithm to estimate the blur kernel and obtain clean LR image, then, they utilized a sub-network to obtain the SR image from the obtained clean LR image. According to previous advances, we can see that deep CNNs are very suitable to SISR. Motivated by that fact, the deep CNN is used in SISR in this paper. 

\subsection{Asymmetric convolution}
Improving the execution speed of a SR model is extremely important for real digital devices \cite{li2019fast,tian2020coarse}. Specifically, decomposing a big convolutional kernel into several small convolutional kernels is useful to accelerate the training speed for computer vision tasks. For instance, Li et al. \cite{li2019fast}
proposed a novel fast spatio-temporal residual block spatio-temporal residual network (FSTRN) in video super-resolution, which is the first work to adopt spatio-temporal 3D convolutions and a cross-space residual learning for the video SR task,  obviously enhancing the performance while keeping a low computational load in contrast with state-of-the-art methods.
Based this idea, asymmetric convolutions are developed. The asymmetric convolutions were applied to approximately represent an existing square-kernel convolutional layer for saving the sum of parameters and boosting the execution speed of target task. The asymmetric convolutions have two different forms: a sequence and element in deep networks \cite{ding2019acnet}.

For the first method, a standard convolution of $m \times m$ can be equivalently converted into a sequence of two layers: a layer of $m \times 1$ and a layer of $1 \times m$ to compress the network and reduce the complexity of the network. These methods broke the rule: two-dimensional convolution with rank of one can be transformed into two one-dimensional convolutions. However, the obtained kernels from deep networks have distributed eigenvalues, where the intrinsic rank is greater than actual value. Also, the transformation process from 2D kernel to 1D kernels would loss information \cite{jin2014flattened}. 

The second method used asymmetric convolutions as elements to design deep CNN. For instance, the Inception-v3 \cite{szegedy2016rethinking} exploited convolution kernels of $1 \times 7$ and $7 \times 1$ rather than a convolution kernel of $7 \times 7$ to reduce the parameters for image recognition. It should be noticed that the similar equivalence was not very effective in low-level task \cite{ding2019acnet}. An efficient dense module with asymmetric convolution (EDANet) \cite{lo2019efficient} method factorized a convolution of $3 \times 3$ into convolutions of $1 \times 3$ and $3 \times 1$ to reduce the computation cost. However, it suffered from performance degradation for semantic segmentation. In terms of dealing with this problem, Ding et al. \cite{ding2019acnet} presented
1D asymmetric convolutions to enhance the features in horizontal and vertical directions, then, gathered their obtained information into the square-kernel layers to guarantee good performance in image recognition. However, there is no work to show the influence of the combination of 1D asymmetric convolutions and square-kernel convolution for low-level vision, especially SISR.
\subsection{Multi-level feature fusion for SISR}
Deep CNNs have shown superior performance for SISR. However, as the growth of depth, the deep architecture may suffer from vanishing or exploding gradients. To resolve this problem, multi-level feature fusion methods were presented. These techniques can be usually classified three kinds: fusion of high-frequency features, fusion of low-frequency features and fusion of high- and low-frequency features.

Fusion of high-frequency features: The methods first used bicubic operation to upsample the LR as input, then, replied on receptive fields of different sizes to extract hierarchical features and fused these features for restoring the HR image. For instance, a deeply-recursive convolutional network (DRCN) \cite{kim2016deeply} referred to recursive operations to construct inference network for SISR. Moreover, a deep recursive residual network (DRRN) \cite{tai2017image} introduced multi-supervision via recursive learning and RL to fuse global and local features for easing the difficulty of training. However, these methods depended on interpolating the LR images to obtain the observed size for training a SR model, which can loss more related information of low-frequency features and take great computational cost. 

Fusion of low-frequency features: To boost the training efficiency and decrease the memory consumption, using the LR image as input and obtaining the HR image at the final layer of network was developed in SISR. For instance, a fast SR convolutional neural network (FSRCNN) only utilized deconvolution layer at the end of the network to learn a non-linear mapping from the low-resolution image to the high-quality image. Also, this method adopted smaller filter size to decrease the sum of parameters for training a SR model, which had good effect on real application \cite{dong2016accelerating}. Alternatively, Lai et al. \cite{lai2017deep} applied Laplacian pyramid and residual learning techniques to progressively infer high-frequency features. Although these methods enjoy fast execution speed and less computational cost, they did not make full use of high-frequency features. That may result in the consequence that the training process was not stable.  

Fusion of high- and low-frequency features: An information distillation network (IDN) \cite{hui2018fast} applied group convolutions to extract richer LR features, and gathered them by the RL technique. Subsequently, the IDN introduced the convolution of $1 \times 1$ to distill extracted low-frequency features. Finally, it used deconvolution and bicubic operation to fuse obtained high-frequency features and construct the HR image. Besides, a deep network with component learning (DNCL) \cite{xie2018fast} used sparse coding to obtain two different components as inputs of the SR network. Specifically, one with low-frequency information was directly used to learn a mapping from the LR image to HR image and obtain the SR features. The other was converted into high-frequency features by upscaling operation. Finally, the DNCL utilized the RL technique to merge obtained high-frequency features and reconstruct the clear image. Moreover, cascaded networks can gather well low-frequency features to boost the robustness of obtained features and stability of training a SR model. For instance, a cascading residual network used multi residual blocks to extract more robust low-level features. After that, it used sub-pixel convolution layer \cite{shi2016real} to transform obtained low-level features into high-level features, and learned more high-level features by cascading a sub-network \cite{hui2019lightweight}. This method not only had fast execution speed, but also achieved superior performance in SISR.

According to previous analysis, it is obvious that fusion of high- and low-frequency features is very competitive to SISR. Thereby, this idea is also used in our designed network to restore the HR image. 
\section{Proposed method}
In this section, we introduce a SR model as well as ACNet in details. First, we illustrate the overall framework in Figs.1 and 2. Then, the components (i.e., an AB, an MEB and an HFFEB) of the ACNet are shown.
\begin{figure*}[!htbp]
\centering
\subfloat{\includegraphics[width=7in]{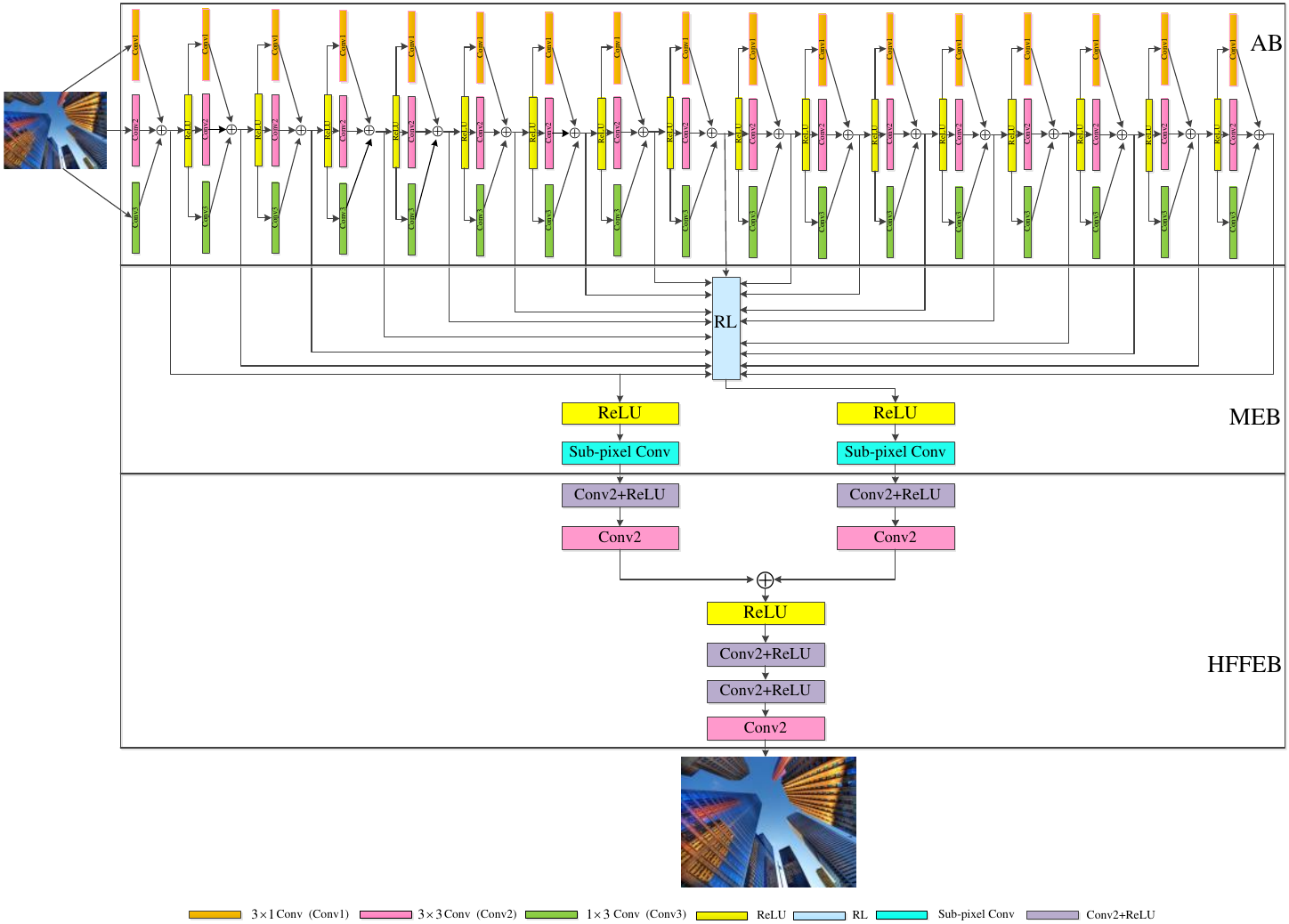}
\label{Fig:a}}
\hfil
\caption{Network architecture of the proposed ACNet.}
\end{figure*}
\begin{figure}[!htbp]
\centering
\subfloat{\includegraphics[width=2.0in]{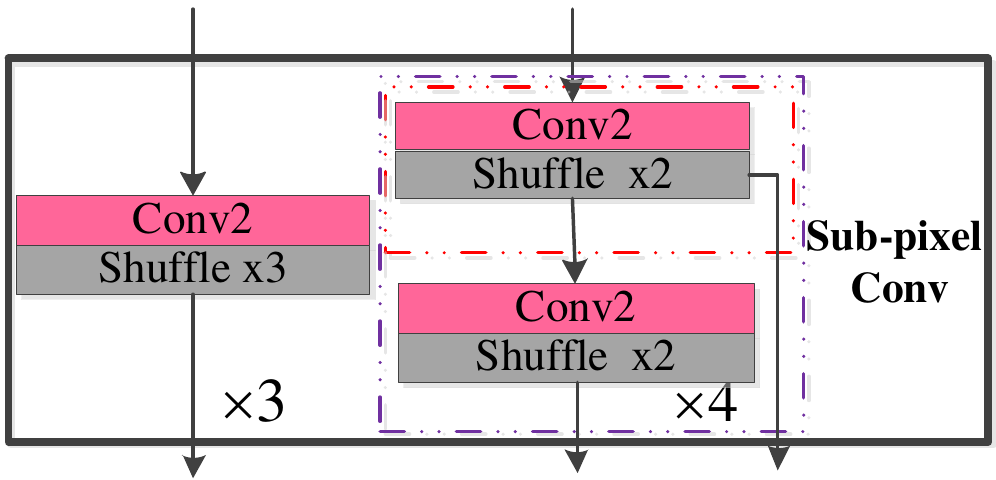}
\label{Fig:a}}
\hfil
\caption{Implementations of the sub-pixel conv.}
\end{figure}
\subsection{Network architecture}
The proposed 23-layer ACNet comprises three blocks: a 17-layer AB, a 1-layer MEB and a 5-layer HFFEB. The AB uses one-dimensional asymmetric convolutions to enhance the effects of local power feature points for improving the expressive ability of the SR model. Also, it can boost the training efficiency and reduce the computational burden. Subsequently, the MEB gathers all hierarchical low-frequency features via the residual learning technique to handle the long-term dependency problem. Moreover, it converts the extracted low-frequency features into high-frequency features. After that, the HFFEB merges the global features from the LR input and the
high-frequency features to learn more accurate super-resolution features. It also addresses excessive feature enhancement problem as well. Additionally, it is utilized to reconstruct a HR image. To vividly express the implementations of ACNet, some terms are defined as follows. Let $I_{LR}$ and $I_{SR}$ denote the LR input and the predicted SR image, respectively, and $f_{AB}$, $f_{MEB}$, $f_{HFFEB}$ and $f_{ACNet}$ denote the corresponding functions of the AB, MEB, HFFEB and the ACNet, respectively. We formulate the SR operation as 
\begin{equation}
\begin{array}{ll}
{I_{SR}} & =   {f_{HFFEB}}({f_{MEB}}({f_{AB}}({I_{LR}})))\\
& = {f_{ACNet}}({I_{LR}}).
\end{array}
\end{equation}

\subsection{Loss function}
We adopt mean square error (MSE) \cite{douillard1995iterative} as the loss function. The MSE loss can be used to minimize the difference between the predicted SR image and the given HR image to train the ACNet model for SISR. The procedure is expressed in (2). 
\begin{equation}
    l(\gamma ) = \frac{1}{{2S}}\sum\limits_{j = 1}^S {\left\| {{f_{ACNet}}(I_{LR}^j) - I_{HR}^j} \right\|} _2^2,
\end{equation}
where $l$ and $\gamma$ represent the loss function and the learned parameters of ACNet, respectively, $I_{LR}^j$ and $I_{HR}^j$ denote the $j$-th LR and HR images, respectively, and $S$ denotes the sum of LR images.
\subsection{Asymmetric block}
It is known that bright colors and complex background may hide some detailed information for an inverse problem. To address this problem, a 17-layer asymmetric block (AB) is proposed to enlarge the effects of salient features at the lowest cost. AB utilizes one-dimensional asymmetric convolutions to intensify the square convolution kernels in the horizontal and vertical directions for improving the influences of local power feature points, which also accelerates the training for a SR model. AB involves four types of layers: Conv1, Conv2, Conv3, and rectified linear unit (ReLU) \cite{krizhevsky2012imagenet}, where Conv1, Conv2, and Conv3 denote the convolutions of sizes $3 \times 1$, $3 \times 3$ and $1 \times 3$, respectively. Conv1 and Conv3 are also treated as one-dimensional asymmetric convolutions, which affects Conv2 via residue learning to enrich the feature space for promoting the expressive ability of the SR model. Conv2 is treated as a square convolution kernel. The ReLU  activation function  is used to non-linearly convert the obtained features. Further, for the first layer, the  channel numbers of the input and output are 3 and 64, respectively. For the second to the seventeenth layers, the numbers of channels of the input and output are 64 and 64, respectively.  For clarity, we define some symbols below. Let
$C_1$, $C_2$, $C_3$ and $R$ denote the corresponding functions of Conv1, Conv2, Conv3, and ReLU, respectively, $O_i^C$ and $O_i^R$ denote the outputs of convolutions and the ReLU of the $i$-th layer in AB, respectively. The above procedures are formulated as follows.
\begin{equation}
   O_i^C = \left\{ {\begin{array}{*{1}{l}}
{{C_1}({I_{LR}}) + {C_2}{\rm{(}}{I_{LR}}{\rm{) + }}{C_{\rm{3}}}({I_{LR}}){\rm{          }}\quad i = 1}\\
\\{{C_1}(O_{i - 1}^R) + {C_2}{\rm{(}}O_{i - 1}^R{\rm{) + }}{C_{\rm{3}}}(O_{i - 1}^R){\rm{  }}\quad i = 2,3,..,17},
\end{array}} \right.
\end{equation}
\begin{equation}
   O_i^R = R(O_i^C){\rm{ }}\quad i = 1,2,...,17,
\end{equation}
where $O_{17}^R$ is the output of AB which is used as the input of the MEB. The '+' indicates $\oplus$ and residues learning in Fig.1. 
\subsection{Memory enhancement block}
Note, increasing the depth of a network is used to weaken the memory abilities from the shallow layers for low-level vision tasks \cite{tai2017memnet}. To solve this problem, a 1-layer memory enhancement block (MEB) is devised. MEB has three steps: the first step merges all hierarchical low-frequency features from AB via residues learning, then, it uses the ReLU to transform the extracted features into non-linearity. This process in (5) can handle long-term dependency problem.
\begin{equation}
O_{MEB}^1 = R(\sum\limits_{i = 1}^{17} {O_i^C)},
\end{equation}
where $O_{MEB}^1$ denotes the output of the first step and is utilized as the input of the second step. The second step converts the obtained low-frequency features into high-frequency features through the sub-pixel convolutional layer \cite{ahn2018fast,tian2020lightweight} as follows. 
\begin{equation}
O_{MEB}^2 = S(O_{MEB}^1),
\end{equation}
where, as shown in Fig. 2, $S$ represents the function of the sub-pixel convolution with a size of $64 \times 3 \times 3 \times 64$, where $64$ is the numbers of input channels and output channels are $64$, and the size of convolutional kernel is $3 \times 3$. Besides, it is noteworthy that the Sub-pixel Conv is implemented by two plugins such as `Conv2+Shuffle $\times2$' and `Conv2+Shuffle $\times3$' for SISR, blind SISR and blind SISR of blind noise, where  `Conv2+Shuffle $\times2$' and `Conv2+Shuffle $\times3$' denote a convolution of $3\times 3$ densely acts Shuffle $\times2$ and Shuffle $\times3$, respectively. Moreover, a `Conv2+Shuffle $\times2$', a `Conv2+Shuffle $\times3$' and two `Conv2+Shuffle $\times2$' is used for $\times2$, $\times3$ and $\times4$, respectively.
Additionally, when a SR model is trained for SISR of certain scale factor, one of Conv2+Shuffle $\times2$', `Conv2+Shuffle $\times3$' and two `Conv2+Shuffle $\times2$' is only used. Otherwise, three modes are simultaneously used for blind SISR and blind SISR of blind noise. 
The $O_{MEB}^2$   indicates the output of the second step. The third step of MEB aims to avoid the loss of global input information. It utilizes sub-pixel convolutions to magnify the output of the first layer in AB, which is complementary to the second step. The process is expressed in into (7).
\begin{equation}
O_{MEB}^3 = S(O_1^R),
\end{equation}
where $O_{MEB}^3$ is the output of the third step. Additionally, $O_{MEB}^3$ together with $O_{MEB}^2$ are used as the input of HFFEB.
\subsection{High-frequency feature enhancement block}
According to Section II. C, it is known that combining the high- and low-frequency features is very useful to stimulate more robust SR features and boost the stability of training. Based on the fact, we propose a 5-layer high-frequency feature enhancement block (HFFEB) to bridge the gap between the obtained HR image and given HR image. HFFEB consists of three types: Conv2+ReLU, Conv2 and ReLU, where Conv2+ReLU means the Conv2 closely connects the ReLU. The three types serve two phases: the first phase fuses the high- and low-frequency features via the 2-layer dual paths of HFFEB and residues learning to provide complementary low-frequency information and improve
the training stability caused by sudden amplification operation of sub-pixel convolutional technique. Specifically, Conv2+ReLU and Conv2 are used in the 19th and 20th layers, respectively, where  their input and output channel numbers are 64. The filter size is $3 \times 3$. ReLU is used to non-linearly transform the extracted high-frequency features by fusing the dual paths. The whole process is expressed as follows:
\begin{equation}
  O_{HFFEB}^1 = R({C_2}(R({C_2}(O_{MEB}^2))) + {C_2}(R({C_2}(O_{MEB}^3)))),
\end{equation}
where $O_{HFFEB}^1$ stands for the output of the first phase in HFFEB.

The second phase is used to prevent excessive feature enhancement problem, so as to obtain more robust SR features and reconstruct the predicted SR image. It has two types of operations: Conv2+ReLU and Conv2. The Conv2+ReLU with a sizes of $64 \times 3 \times 3 \times 64$ is utilized in the 21th and 22th layers of ACNet, where the input and output channels numbers are 64 and Conv2's filiter size is $3 \times3$. The final layer 
of ACNet only involve a Conv2 with a size of $64 \times 3 \times 3 \times 3$ to reconstruct the HR image, where the input and output channel numbers are $64$ and $3$, respectively. The process can be expressed as
\begin{equation}
  {I_{SR}} = {C_2}(R({C_2}(R({C_2}(O_{HFFEB}^1))))).
\end{equation}
\section{Experiments}
\subsection{Training dataset}
 Existing methods such as \cite{ahn2018fast,lim2017enhanced} used benchmark dataset DIV2K dataset \cite{timofte2017ntire} to train a SR model. To make the experiments fair, we choose the high-quality DIV2K dataset as our training dataset. The DIV2K dataset is composed of three parts: 800 training images, 100 validation images, and 100 test images. Our experiments are conducted for different scale factors, including $\times 2$, $\times 3$ and $\times 4$. It is known that differences in textures and edges of different LR images have great influence on SR model. To address this problem, image augmentation has obtained good performance in image \cite{krizhevsky2012imagenet,tian2020designing} and video applications \cite{yuan2020visual}. Based on this idea, a two-step mechanism \cite{tian2020coarse,tian2020lightweight} is used to enlarge the training dataset for improving the generalization ability of the SR model. The first step combines the training and validation datasets as a novel training dataset. That is, 100  validation images (i.e., LR and HR images) of different scale factors (i.e., $\times 2$, $\times 3$, $\times 4$) are extended to the corresponding scale training dataset of DIV2K. The second step uses random horizontal flips and $90^\circ$ rotation operations to further augment the training dataset. Similarly, we also use this approach to enlarge the training dataset in our ACNet. Besides, to accelerate the training speed, each LR image is cropped into patches of $64 \times 64$ as the input of the ACNet. 
\subsection{Testing datasets}
Four benchmark datasets including Set5 \cite{bevilacqua2012low}, Set14 \cite{yang2010image}, BSD100 \cite{martin2001database} and Urban100 \cite{huang2015single} are used as test datasets in this paper. These datasets have three different scale factors: $ \times 2$, $ \times 3$ and $ \times 4$. Set14 and Set5 contain fourteen and five natural images of different scenes, respectively. BSD100/B100 and Urban100/U100 contain 100 color images of different backgrounds, respectively. 

It should be noted that the SR methods such as VDSR \cite{kim2016accurate} and MemNet \cite{tai2017memnet} used Y channel in the YCbCr space to conduct experiments in general. Thus, the obtained RGB from the ACNet need be converted into the Y channel to verify the SR effect in all the experiments.
\subsection{Implementation details}
ACNet adopts the MSE loss function and the Adam optimizer \cite{douillard1995iterative} with initial beta\_1 of 0.9, beta\_2 of 0.999 and epsilon of ${10^{-8}}$. And training procedure has 660,000 steps. The mini-batch size is set to be 16. The initial learning rate is set as ${10^{-4}}$, which is halved every 400,000 steps. More initial parameter settings can be found in \cite{ahn2018fast}.

The code of ACNet runs on a PC with Ubuntu of 16.04, a CPU of Intel Core i7-7800, a RAM of 16G and a GPU of GTX 1080Ti, where Nvidia CUDA of 9.0 and CuDNN of 7.5 are used to accelerate the GPU for training a SR model. 
\subsection{Network analysis}
ACNet can enhance the effects of local power feature points to improve the performance and accelerate the speed for SISR. It is implemented by three components: an AB, a MEB and a HFFEB. The AB uses one-dimensional asymmetric convolutions to increase the square convolution kernels in horizontal and vertical directions for promoting the super-resolution performance. The MEB gathers all hierarchical low-frequency features from the AB via the RL technique to deal with the long-term dependency issue. Also, it can convert obtained low-frequency information into high-frequency information. The HFFEB merges low- and high-frequency features by two phases to obtain more robust super-resolution features, which can enhance the stability of training process. Additionally, it can prevent excessive feature enhancement problem. 
It is also responsible to predict the high-quality image. The principles of these techniques are shown as follows in details. 
\subsubsection{AB}
It is known that enlarging the receptive field can facilitate more context information to enhance the pixels in image applications \cite{tian2020designing}. Increasing the width is very effective method to expand the receptive field. Most of these methods equally treats all pixel points via different paths to improve the pixels, which enlarges the influence of non-critical feature points and results in low training efficiency \cite{tian2020attention}. Taking into the mentioned factors consideration, a 17-layer AB is proposed. Each layer of the AB utilizes one-dimensional (also called 1D) asymmetric convolutions \cite{ding2019acnet} to intensify the square convolution kernels in horizontal and vertical directions for promoting the super-resolution performance. The one-dimensional asymmetric convolutions are $3 \times 1$ and $1 \times 3$. The square convolution is $3 \times 3$. After that, the ReLU is used to change the collected features into non-linear features. More detailed information of the AB is illustrated in Section III.C. Besides, the proposed AB can be explained in theory and design of network architecture as follows. In theory, according to rank of one from 2D kernel, the 2D kernel can be converted into two 1D asymmetric convolutions. However, learned kernels of deep CNNs have eigenvalues, which has higher rank than that of one in practice. So that, the transformation process will loss information. To guarantee the excellent performance 
of goal task, the asymmetric convolutions are used to consolidate the square kernel. That not only enhances the effect of key features, but also improves the performance in image applications.

In terms of design of the network architecture, the AB observes three principles: intensity of training, SISR performance, and training efficiency. For the first aspect, shallow network architecture is useful to train a model \cite{zhang2017beyond}. Inspired by that, we design a 17-layer network, which does not result in training difficulty of the network. 

For the second aspect, we can see that expanding the width of the network can enlarge the receptive field to capture more detailed information for boosting the SR effect \cite{tian2020designing}. Also, expanding the diversity of the network can obtain more robust features \cite{zhang2018residual}. Motivated by that, the combination of 1D asymmetric convolutions and square convolution used in the AB is proper.  

For the third aspect, we use small filter size (i.e., $3 \times 3$) to decrease sum of parameters. Finally, the one-dimensional convolutions (i.e., $3 \times 1$ and $1 \times 3$) are also superior to improving the training efficiency in SISR. 
\begin{table}[htbp!]
\caption{Average PSNR and SSIM of five methods for scale factor of $\times 2$ on B100 and U100 in SISR.}
\label{tab:1}
\centering
\scalebox{0.9}[0.9]{
\begin{tabular}{|c|c|c|c|c|}
\hline
\multirow{2}{*}{Methods}
&B100 &U100 \\
\cline{2-3} &PSNR/SSIM	&PSNR/SSIM\\
\hline
SB	&30.90/0.9153	&31.73/0.8943\\
\hline
AEB	&30.98/0.9155  &31.78/0.8945\\
\hline
AB &31.07/0.9176    &31.82/0.8951\\
\hline
The combination of AB and MEB  &31.48/0.9219 &31.94/0.8968\\
\hline
ACNet (Ours) &31.79/0.9245	 &32.06/0.8978\\
\hline
\end{tabular}}
\label{tab:booktabs}
\end{table}

To verify the effectiveness of the AB in performance and efficiency, we use SB, AEB and AB as comparative methods to conduct experiments. The SB denotes a 17-layer network, where filter size of each layer is $3 \times 3$. The 17-layer AEB with all filter sizes of $3 \times 3$ has the same architecture as the AB. Specifically, these methods connect a sub-pixel convolution and a convolution in all experiments, where the sub-pixel convolution method be applied to convert the low-resolution image into the high-definite image. A convolution is adopted to predict high-quality image. As shown in Table I, the AB obtained higher peak signal-to-noise ration (PSNR) and structural similarity index (SSIM) \cite{hore2010image} than that of the SB on public datasets, e.g. B100 and U100 under scale factor of $\times 2$, respectively. The AEB performs better than the SB on B100 and U100 for $\times 2$, respectively. These illustrations test that the proposed AB is very useful to SISR. Additionally, the AB achieves great improvements of performance in contrast with the AEB on B100 and U100 for $\times 2$, respectively. This shows that the proposed AB with 1D asymmetric convolutions is very effectiveness to enhance the effects of local power features for SISR.
\begin{table}[htbp!]
\caption{Running time of the AEB and AB for predicting a HR image of sizes of $256 \times 256$, $512 \times 512$,$1024 \times 1024$.}
\label{tab:1}
\centering
\scalebox{1}[1]{
\begin{tabular}{|c|c|c|}
\hline
\multirow{3}{*}{Sizes} &
\multicolumn{2}{c|}{Methods}\\
\cline{2-3}
 &AEB &AB\\
\cline{2-3}
 & \multicolumn{2}{c|}{$\times 2$}\\
\hline
$256 \times 256$ &0.01522	  &0.01512\\
\hline
$512 \times 512$ &0.01771	&0.01763\\
\hline
$1024 \times 1024$ &0.02821	&0.02817\\
\hline
\end{tabular}}
\label{tab:booktabs}
\end{table}

In efficiency of the AB for SISR, we choose run-time of a given figure with  different sizes, parameters and flops as metrics to design some experiments as follows. The AB has fast executive speed than that of the AEB on a given figure with different sizes, such as $256 \times 256$, $512 \times 512$ and $1024 \times 1024$ for $\times 2$ as shown in Table II. The AB enjoys less parameters and flops than that of the AEB as described in Table III. These describe that the proposed AB has higher efficiency in SISR. According to theory analysis, design of the network architecture, and test of performance and efficiency, we can see that the AB is proper and beneficial for SISR.
\begin{table}[htbp!]
\caption{Complexity from the AEB and AB.}
\label{tab:1}
\centering
\scalebox{0.9}[1]{
\begin{tabular}{|c|c|c|}
\hline
Methods &Parameters &Flops\\
\hline
AEB  &1,813K	&7.90G\\
\hline
AB	&1,025K	&4.67G\\
\hline
\end{tabular}}
\label{tab:booktabs}
\end{table}
\subsubsection{MEB}
According to ResNet\cite{he2016deep} and MemNet \cite{tai2017memnet}, we can see that increases the depth of the network will reduce the effects of shallow layers on deep layers, which makes the performance degradation. Additionally, deeper network architecture may enlarge the risk of training difficulty. To resolve this problem, Zhang et al. \cite{zhang2018residual} fused hierarchical low-frequency features to improve the memory ability of shallow layers and facilitate more robust low-frequency features of deep layer, according to nature of SISR and principle of residual learning technique. Inspired by that, we design a 1-layer MEB by three steps. The first step fuses hierarchical low-frequency information from the AB to enhance the influence of shallow layers on deep layers for addressing long-term dependency problem. The second step uses 1-layer sub-pixel convolution in the MEB to map obtained low-frequency information into high-frequency information. The up-sampling operation is set into the deep layer of the network, which can reduce the training cost and make the training process easier. Because the up-sampling operation may loss some information, the third step amplifies global input information by the sub-pixel convolution to offer complementary features for the second step, where the obtained features from the second and third steps can be further handled by the HFFEB. Additionally, Table I proves the effectiveness of the MEB, where the combination of AB and MEB obtains superior performance against the AB on U100 and B100 for scale factor of $\times 2$, respectively. According to these illustrations, it is known that the proposed MEB is rational. 
\subsubsection{HFFEB}
Applying the bicubic interpolation operation to up-scale LR image as input for training a SR model can result in greater computational cost \cite{dong2016accelerating}. However, utilizing the given low-resolution image as input and up-sampling operation as the final layer to train the SR model losses some low-frequency information caused by sudden shock from up-sampling operation, which obtains poor performance. To resolve this problem, an extra module was used to refine the HR features and obtain more robust HR features \cite{ahn2018image}.
Inspired by that fact, the HFFEB is proposed. It is implemented by two phases. The first phase fuses learned high-frequency features via the RL technique from the MEB and low-frequency features from global input to enhance the robustness of obtained SR features. It is noted that the feature enhancement of the first phase may make the pixels excessive enhancement. Thus, the second phase uses 2-layer Conv2+ReLU to learn more accurate SR features, which can arrow
disparities between the predicted high-quality image image and the given high-resolution image. Also, it is also useful to improve the training stability. Finally, a convolution is applied to reconstruct a SR image. Further, the good performance of the mentioned process is tested in Table I. Specifically, due to different shooting  environment and devices, the texture, edge and detailed information from the captured images are discrepant. Based on this reason, a robust SR model is very important for different conditions in the real world. In terms of this factor above, although our ACNet improves PSNR a little that than of 
other popular methods in certain condition, it has stable performance for all conditions, which is very useful to real digital devices. 
\subsection{Comparisons with state-of-the-arts}
To comprehensively evaluate the SISR and blind SISR performance of the proposed ACNet, we use quantitative and qualitative analysis to conduct experiments. The quantitative analysis includes SR results (i.e., PSNR and SSIM), perceptual quality of feature similarity index (FSIM) \cite{zhang2011fsim}, run-time of a predicted HR image and complexity of some state-of-the-arts, such as Bicubic, A+ \cite{timofte2014a+}, jointly optimized regressors (JOR) \cite{dai2015jointly}, RFL \cite{schulter2015fast}, self-exemplars super-resolution (SelfEx) \cite{huang2015single}, the cascade of sparse coding based networks (CSCN) \cite{wang2015deep}, residual encoder-decoder network (RED) \cite{mao2016image}, a denoising CNN (DnCNN) \cite{zhang2017beyond}, trainable nonlinear reaction diffusion (TNRD) \cite{chen2016trainable}, fast dilated residual super-resolution convolution network (FDSR) \cite{lu2018fast}, SRCNN \cite{dong2014learning}, FSRCNN \cite{dong2016accelerating}, residue context sub-network (RCN) \cite{shi2017structure}, VDSR \cite{kim2016accurate}, DRCN \cite{kim2016deeply}, context-wise network fusion (CNF) \cite{ren2017image}, Laplacian super-resolution network (LapSRN) \cite{lai2017deep}, DRRN \cite{tai2017image}, MemNet \cite{tai2017memnet}, CARN-M \cite{ahn2018fast}, wavelet domain residual network (WaveResNet) \cite{bae2017beyond}, convolutional principal component analysis (CPCA) \cite{xu2018self}, deep recurrent fusion network (DRFN) \cite{yang2018drfn}, wiener filter in similarity domain SR (WSDSR) \cite{cruz2017single}, deep network with component learning (DNCL) \cite{xie2018fast}, new architecture of deep recursive convolutional networks for SR (NDRCN)\cite{cao2019new} and 
 LESRCNN \cite{tian2020lightweight}
on four pubic datasets (e.g. Set5, Set14, B100 and U100) for different scale factors of $\times 2$, $\times 3$ and $\times 4$, respectively. The qualitative analysis uses some visual figures (i.e., SR image,  Y channel of SR image, error image and edge image) from different aspects, i.e., flat areas, textures, fine details and edge information to verify the SR results of different methods SISR, where the error image denotes the difference between the predicted SR and the given high-quality image. Additionally, we extend the ACNet model for blind SISR of blind noise task, which is comprehensively shown in the final part of this sub-section.
\begin{table}[htbp!]
\caption{Average PSNR/SSIM values of different methods for three scale factors of $\times 2$, $\times 3$ and $\times 4$ on the Set5.}
\label{tab:1}
\centering
\scalebox{0.90}[0.90]{
\begin{tabular}{|c|c|c|c|c|}
\hline
\multirow{2}{*}{Dataset} &
\multirow{2}{*}{Methods} &
$\times 2$ & $\times 3$ & $\times 4$\\
\cline{3-5} & &PSNR/SSIM &PSNR/SSIM &PSNR/SSIM\\
\hline
\multirow{18}{*}{Set5} &
Bicubic	&33.66/0.9299	&30.39/0.8682	&28.42/0.8104\\
\cline{2-5} &
A+\cite{timofte2014a+}	&36.54/0.9544	&32.58/0.9088	&30.28/0.8603\\
\cline{2-5} &
JOR\cite{dai2015jointly}	&36.58/0.9543	&32.55/0.9067	&30.19/0.8563\\
\cline{2-5} &
RFL \cite{schulter2015fast}	&36.54/0.9537	&32.43/0.9057	&30.14/0.8548\\
\cline{2-5} &
SelfEx\cite{huang2015single}	&36.49/0.9537 &32.58/0.9093 &30.31/0.8619\\
\cline{2-5} &
CSCN\cite{wang2015deep} &36.93/0.9552 &33.10/0.9144 &30.86/0.8732\\
\cline{2-5} &
RED\cite{mao2016image} &37.56/0.9595 &33.70/0.9222	&31.33/0.8847\\
\cline{2-5} &
DnCNN\cite{zhang2017beyond}	&37.58/0.9590	&33.75/0.9222 &31.40/0.8845\\
\cline{2-5} &
TNRD\cite{chen2016trainable} &36.86/0.9556	&33.18/0.9152 &30.85/0.8732\\
\cline{2-5} &
FDSR\cite{lu2018fast} &37.40/0.9513	&33.68/0.9096 &31.28/0.8658\\
\cline{2-5} &
SRCNN\cite{dong2014learning}	&36.66/0.9542	&32.75/0.9090 &30.48/0.8628\\
\cline{2-5} &
FSRCNN\cite{dong2016accelerating} &37.00/0.9558	&33.16/0.9140 &30.71/0.8657\\
\cline{2-5} &
RCN\cite{shi2017structure} &37.17/0.9583	&33.45/0.9175	&31.11/0.8736\\
\cline{2-5} &
VDSR\cite{kim2016accurate} &37.53/0.9587	&33.66/0.9213	&31.35/0.8838\\
\cline{2-5} &
DRCN\cite{kim2016deeply} &37.63/0.9588	&33.82/0.9226	&31.53/0.8854\\
\cline{2-5} &
CNF\cite{ren2017image}	&37.66/0.9590	&33.74/0.9226	&31.55/0.8856\\
\cline{2-5} &
LapSRN\cite{lai2017deep} &37.52/0.9590	&-	&31.54/0.8850\\
\cline{2-5} &
DRRN\cite{tai2017image} &\textcolor{blue}{37.74}/0.9591  &34.03/0.9244 &31.68/0.8888 \\
\cline{2-5} &
MemNet\cite{tai2017memnet}  &\textcolor{red}{37.78}/0.9597  &\textcolor{blue}{34.09}/\textcolor{red}{0.9248} &31.74/0.8893\\
\cline{2-5} &
CARN-M\cite{ahn2018fast}  &37.53/0.9583 &33.99/0.9236 &\textcolor{red}{31.92/0.8903}\\
\cline{2-5} &
WaveResNet\cite{bae2017beyond} &37.57/0.9586  &33.86/0.9228 &31.52/0.8864\\
\cline{2-5} &
CPCA\cite{xu2018self} &34.99/0.9469 &31.09/0.8975 &28.67/0.8434\\
\cline{2-5} &
DRFN\cite{yang2018drfn} &37.71/\textcolor{blue}{0.9595} &34.01/0.9234 &31.55/0.8861\\
\cline{2-5} &
WSDSR\cite{cruz2017single}	&37.16/0.9583	&33.45/0.9196	&31.29/0.8821\\
\cline{2-5} &
DNCL\cite{xie2018fast}	&37.65/\textcolor{red}{0.9599}	&33.95/0.9232	&31.66/0.8871\\
\cline{2-5} &
NDRCN\cite{cao2019new} &37.73/0.9596 &33.90/0.9235 &31.50/0.8859\\
\cline{2-5} &
LESRCNN\cite{tian2020lightweight} 	&37.65/0.9586	&33.93/0.9231	&\textcolor{blue}{31.88}/\textcolor{red}{0.8903}\\
\cline{2-5} &
ACNet (Ours)	& 37.72/0.9588	&\textcolor{red}{34.14}/\textcolor{blue}{0.9247}	&31.83/\textcolor{red}{0.8903}\\
\cline{2-5} &
ACNet-B (Ours) &37.60/0.9584 &34.07/0.9243 &31.82/0.8901\\
\hline
\end{tabular}}
\label{tab:booktabs}
\end{table}
\begin{table}[t!]
\caption{Average PSNR/SSIM values of different methods for three scale factors of $\times 2$, $\times 3$ and $\times 4$ on the Set14.}
\label{tab:1}
\centering
\scalebox{0.90}[0.90]{
\begin{tabular}{|c|c|c|c|c|}
\hline
\multirow{2}{*}{Dataset} &
\multirow{2}{*}{Methods} &
$\times 2$ & $\times 3$ & $\times 4$\\
\cline{3-5} & &PSNR/SSIM &PSNR/SSIM &PSNR/SSIM\\
\hline
\multirow{20}{*}{Set14} &
Bicubic	&30.24/0.8688	&27.55/0.7742	&26.00/0.7027\\
\cline{2-5} &
A+\cite{timofte2014a+}	&32.28/0.9056	&29.13/0.8188	&27.32/0.7491\\
\cline{2-5} &
JOR\cite{dai2015jointly}	&32.38/0.9063	&29.19/0.8204	&27.27/0.7479\\
\cline{2-5} &
RFL\cite{schulter2015fast}	&32.26/0.9040	&29.05/0.8164	&27.24/0.7451\\
\cline{2-5} &
SelfEx\cite{huang2015single}	&32.22/0.9034 &29.16/0.8196	&27.40/0.7518\\
\cline{2-5} &
CSCN\cite{wang2015deep} &32.56/0.9074	&29.41/0.8238	&27.64/0.7578\\
\cline{2-5} &
RED\cite{mao2016image} &32.81/0.9135	&29.50/0.8334	&27.72/0.7698\\
\cline{2-5} &
DnCNN\cite{zhang2017beyond} &33.03/0.9128	&29.81/0.8321	&28.04/0.7672\\
\cline{2-5} &
TNRD\cite{chen2016trainable} &32.51/0.9069	&29.43/0.8232	&27.66/0.7563\\
\cline{2-5} &
FDSR\cite{lu2018fast} &33.00/0.9042	&29.61/0.8179	&27.86/0.7500\\
\cline{2-5} &
SRCNN\cite{dong2014learning} &32.42/0.9063 &29.28/0.8209 &27.49/0.7503\\
\cline{2-5} &
FSRCNN\cite{dong2016accelerating} &32.63/0.9088 &29.43/0.8242	&27.59/0.7535\\
\cline{2-5} &
RCN\cite{shi2017structure} &32.77/0.9109	&29.63/0.8269	&27.79/0.7594\\
\cline{2-5} &
VDSR\cite{kim2016accurate} &33.03/0.9124	&29.77/0.8314	&28.01/0.7674\\
\cline{2-5} &
DRCN\cite{kim2016deeply} &33.04/0.9118	&29.76/0.8311	&28.02/0.7670\\
\cline{2-5} &
CNF\cite{ren2017image} &\textcolor{blue}{33.38}/0.9136 &29.90/0.8322 &28.15/0.7680\\
\cline{2-5} &
LapSRN\cite{lai2017deep}	&33.08/0.9130	&29.63/0.8269 &28.19/0.7720\\
\cline{2-5} &
IDN\cite{hui2018fast} &33.30/0.9148 &29.99/0.8354 &28.25/0.7730\\
\cline{2-5} &
DRRN\cite{tai2017image} &33.23/0.9136	&29.96/0.8349	&28.21/0.7720\\
\cline{2-5} &
BTSRN\cite{fan2017balanced} &33.20/- &29.90/- &28.20/-\\
\cline{2-5} &
MemNet\cite{tai2017memnet}	&33.28/0.9142	&30.00/0.8350	&28.26/0.7723\\
\cline{2-5} &
CARN-M\cite{ahn2018fast} &33.26/0.9141	&30.08/0.8367	&\textcolor{blue}{28.42}/0.7762\\
\cline{2-5} &
WaveResNet\cite{bae2017beyond} &33.09/ 0.9129 &29.88/0.8331 &28.11/0.7699\\
\cline{2-5} &
CPCA\cite{xu2018self} &31.04/0.8951  &27.89/0.8038 &26.10/0.7296\\
\cline{2-5} &
DRFN\cite{yang2018drfn} &33.29/0.9142 &30.06/0.8366 &28.30/0.7737\\
\cline{2-5} &
WSDSR\cite{cruz2017single}	&32.57/0.9108	&29.39/0.8302	&27.59/0.7659\\
\cline{2-5} &
DNCL\cite{xie2018fast} &33.18/0.9141	&29.93/0.8340	&28.23/0.7717\\
\cline{2-5} &
NDRCN\cite{cao2019new} &33.20/0.9141 &29.88/0.8333 &28.10/0.7697\\
\cline{2-5} &
LESRCNN\cite{tian2020lightweight}	&33.32/0.9148	&30.12/0.8380	&28.44/0.7772\\
\cline{2-5} &
ACNet (Ours)	&\textcolor{red}{33.41/0.9160}	&\textcolor{red}{30.19/0.8398}	&\textcolor{red}{28.46/0.7788}\\
\cline{2-5} &
ACNet-B (Ours) &33.32/\textcolor{blue}{0.9151} &\textcolor{blue}{30.15/0.8386} &28.41/\textcolor{blue}{0.7773}\\
\hline
\end{tabular}}
\label{tab:booktabs}
\end{table}

Quantitative analysis: The PSNR and SSIM values of different SR methods are shown in Tables IV-VII. In this Table IV, we can see that the proposed ACNet obtains higher PSNR and SSIM than that of the other methods for $\times 3$ on the Set5, where the red and blue lines express the best and second results for SISR, respectively. And it is close to the best results for $\times 2$ and $\times 4$, respectively. The proposed ACNet obtains superior performance than that of other popular methods for three scale factors (i.e.,  $\times 2$, $\times 3$ and $\times 4$) on the Set14 as shown in Table V.  For 
instance, the ACNet achieves the improvements in PSNR value of 0.11dB and SSIM value of 0.0031 than that of the CARN-M for scale factor of $\times 3$ on the Set14. Additionally, the ACNet obtains comparative SR results on large-scale datasets, i.e., B100 and U100. From the Tables VI and VII, it is known that the proposed ACNet has obvious improvements that of state-of-the-art approaches, such as WSDSR. For example, the ACNet has improvement of 0.57dB and 0.0064 than that of the WSDSR for PSNR and SSIM on the B100 under scale factor of $\times 2$, respectively. The ACNet has achieved gain both of PSNR of 0.43dB and SSIM of 0.1368 in contrast to the MemNet for $\times 4$ on the U100 in Table VII. Besides, the blind SR model of ACNet (also regarded as ACNet-B) trained by single model for varying scales (i.e., $\times 2$, $\times 3$ and $\times 4$) also obtains the same great performance as ACNet for $\times 2$, $\times 3$ and $\times 4$ on four public datasets, i.e., Set5, Set14, B100 and U100 in SISR, respectively. For instance, the ACNet-B has obtained improvements of 0.26dB in PSNR and 0.0058 in SSIM than that of the DNCL for $\times 3$ on the Set14. These show that ACNet-B is very useful to recover low-resolution images of different conditions in the real-world. Specifically, in Tables V, VI and VII, the best and second both of PSNR and SSIM are denoted as red line and blue line, respectively. 

\begin{table}[t!]
\caption{Average PSNR/SSIM values of different methods for three scale factors of $\times 2$, $\times 3$ and $\times 4$ on the B100.}
\label{tab:1}
\centering
\scalebox{0.90}[0.90]{
\begin{tabular}{|c|c|c|c|c|}
\hline
\multirow{2}{*}{Dataset} &
\multirow{2}{*}{Methods} &
$\times 2$ & $\times 3$ & $\times 4$\\
\cline{3-5} & &PSNR/SSIM &PSNR/SSIM &PSNR/SSIM\\
\hline
\multirow{17}{*}{B100} &
Bicubic	&29.56/0.8431	&27.21/0.7385	&25.96/0.6675\\
\cline{2-5} &
A+\cite{timofte2014a+}	&31.21/0.8863	&28.29/0.7835	&26.82/0.7087\\
\cline{2-5} &
JOR\cite{dai2015jointly}	&31.22/0.8867	&28.27/0.7837 &26.79/0.7083\\
\cline{2-5} &
RFL\cite{schulter2015fast} &31.16/0.8840	&28.22/0.7806	&26.75/0.7054\\
\cline{2-5} &
SelfEx\cite{huang2015single}	&31.18/0.8855	&28.29/0.7840 &26.84/0.7106\\
\cline{2-5} &
CSCN\cite{wang2015deep} &31.40/0.8884	&28.50/0.7885 &27.03/0.7161\\
\cline{2-5} &
RED\cite{mao2016image} &31.96/0.8972	&28.88/0.7993 &27.35/0.7276\\
\cline{2-5} &
DnCNN\cite{zhang2017beyond}	&31.90/0.8961	&28.85/0.7981	&27.29/0.7253\\
\cline{2-5} &
TNRD\cite{chen2016trainable} &31.40/0.8878	&28.50/0.7881	&27.00/0.7140\\
\cline{2-5} &
FDSR\cite{lu2018fast} &31.87/0.8847	&28.82/0.7797	&27.31/0.7031\\
\cline{2-5} &
SRCNN\cite{dong2014learning}	&31.36/0.8879	&28.41/0.7863	&26.90/0.7101\\
\cline{2-5} &
FSRCNN\cite{dong2016accelerating} &31.53/0.8920	&28.53/0.7910	&26.98/0.7150\\
\cline{2-5} &
VDSR\cite{kim2016accurate} &31.90/0.8960	&28.82/0.7976	&27.29/0.7251\\
\cline{2-5} &
DRCN\cite{kim2016deeply} &31.85/0.8942	&28.80/0.7963	&27.23/0.7233\\
\cline{2-5} &
CNF\cite{ren2017image}	&31.91/0.8962	&28.82/0.7980	&27.32/0.7253\\
\cline{2-5} &
LapSRN\cite{lai2017deep}	&31.80/0.8950 &-	&27.32/0.7280\\
\cline{2-5} &
IDN\cite{hui2018fast} &32.08/0.8985 &28.95/0.8013 &27.41/0.7297\\
\cline{2-5} &
DRRN\cite{tai2017image} &\textcolor{blue}{32.05}/0.8973 &28.95/0.8004 &27.38/0.7284\\
\cline{2-5} &
BTSRN\cite{fan2017balanced} &\textcolor{blue}{32.05}/-    &\textcolor{blue}{28.97}/- &27.47/- \\
\cline{2-5} &
CARN-M\cite{ahn2018fast}	&31.92/0.8960	&28.91/0.8000	&27.44/0.7304\\
\cline{2-5} &
DRFN\cite{yang2018drfn} &32.02/\textcolor{red}{0.8979} &28.93/0.8010  &27.39/0.7293\\
\cline{2-5} &
WSDSR\cite{cruz2017single} &31.49/0.8914 &28.59/0.7934 &27.12/0.7215\\
\cline{2-5} &
DNCL\cite{xie2018fast} &31.97/0.8971	&28.91/0.7995	&27.39/0.7282\\
\cline{2-5} &
NDRCN\cite{cao2019new}  &32.00/0.8975 &28.86/0.7991 &27.30/0.7263\\
\cline{2-5} &
LESRCNN\cite{tian2020lightweight}	&31.95/0.8964	&28.91/0.8005	&27.45/0.7313\\
\cline{2-5} &
ACNet (Ours)	&\textcolor{red}{32.06}/\textcolor{blue}{0.8978}	&\textcolor{red}{28.98/0.8023}	&\textcolor{red}{27.48/0.7326}\\
\cline{2-5} &
ACNet-B (Ours) &31.97/0.8970 &\textcolor{blue}{28.97/0.8016} &\textcolor{blue}{27.46/0.7316}\\ 
\hline
\end{tabular}}
\label{tab:booktabs}
\end{table}
\begin{table}[t!]
\caption{Average PSNR/SSIM values of different methods for three scale factors of $\times 2$, $\times 3$ and $\times 4$ on the U100.}
\label{tab:1}
\centering
\scalebox{0.90}[0.90]{
\begin{tabular}{|c|c|c|c|c|}
\hline
\multirow{2}{*}{Dataset} &
\multirow{2}{*}{Model} &
$\times 2$ & $\times 3$ & $\times 4$\\
\cline{3-5} & &PSNR/SSIM &PSNR/SSIM &PSNR/SSIM\\
\hline
\multirow{18}{*}{U100} &
Bicubic	&26.88/0.8403	&24.46/0.7349	&23.14/0.6577\\
\cline{2-5} &
A+\cite{timofte2014a+}	&29.20/0.8938	&26.03/0.7973	&24.32/0.7183\\
\cline{2-5} &
JOR\cite{dai2015jointly} &29.25/0.8951	&25.97/0.7972	&24.29/0.7181\\
\cline{2-5} &
RFL\cite{schulter2015fast}	&29.11/0.8904	&25.86/0.7900	&24.19/0.7096\\
\cline{2-5} &
SelfEx\cite{huang2015single}	&29.54/0.8967	&26.44/0.8088 &24.79/0.7374\\
\cline{2-5} &
DnCNN\cite{zhang2017beyond} &30.74/0.9139	&27.15/0.8276	&25.20/0.7521\\
\cline{2-5} &
TNRD\cite{chen2016trainable} &29.70/0.8994	&26.42/0.8076	&24.61/0.7291\\
\cline{2-5} &
FDSR\cite{lu2018fast} &30.91/0.9088	&27.23/0.8190	&25.27/0.7417\\
\cline{2-5} &
SRCNN\cite{dong2014learning} &29.50/0.8946	&26.24/0.7989	&24.52/0.7221\\
\cline{2-5} &
FSRCNN\cite{dong2016accelerating} &29.88/0.9020	&26.43/0.8080	&24.62/0.7280\\
\cline{2-5} &
VDSR\cite{kim2016accurate} &30.76/0.9140	&27.14/0.8279	&25.18/0.7524\\
\cline{2-5} &
DRCN\cite{kim2016deeply} &30.75/0.9133	&27.15/0.8276	&25.14/0.7510\\
\cline{2-5} &
LapSRN\cite{lai2017deep} &30.41/0.9100	&-	&25.21/0.7560\\
\cline{2-5} &
IDN\cite{hui2018fast} &31.27/0.9196 &27.42/0.8359 &25.41/0.7632\\
\cline{2-5} &
DRRN\cite{tai2017image} &31.23/0.9188	&27.53/0.8378	&25.44/0.7638\\
\cline{2-5} &
BTSRN\cite{fan2017balanced} &\textcolor{blue}{31.63}/- &27.75/- &25.74- \\
\cline{2-5} &
MemNet\cite{tai2017memnet}	&31.31/0.9195	&27.56/0.8376	&25.50/0.7630\\
\cline{2-5} &
CARN-M\cite{ahn2018fast} &30.83/\textcolor{blue}{0.9233}	&26.86/0.8263	
&25.63/0.7688\\
\cline{2-5} &
WaveResNet\cite{bae2017beyond} &30.96/0.9169 &27.28/0.8334 &25.36/0.7614\\
\cline{2-5} &
CPCA\cite{xu2018self} &28.17/0.8990 &25.61/0.8123 &23.62/0.7257\\
\cline{2-5} &
DRFN\cite{yang2018drfn} &31.08/0.9179 &27.43/0.8359 &25.45/0.7629\\
\cline{2-5} &
WSDS\cite{cruz2017single} &30.23/0.9066	&26.91/0.8204	&25.11/0.7492\\
\cline{2-5} &
DNCL\cite{xie2018fast} &30.89/0.9158	&27.27/0.8326	&25.36/0.7606\\
\cline{2-5} &
NDRCN\cite{cao2019new}  &31.06/0.9175 &25.16/0.7546\\
\cline{2-5} &
LESRCNN\cite{tian2020lightweight}	&31.45/0.9206	&27.70/0.8415	&25.77/0.7732\\
\cline{2-5} &
ACNet (Ours)	&\textcolor{red}{31.79/0.9245}	&\textcolor{red}{27.97/0.8482}	&\textcolor{red}{25.93/0.7798}\\
\cline{2-5} &
ACNet-B (Ours) &31.57/0.9222 &\textcolor{blue}{27.88/0.8447} &\textcolor{blue}{25.86/0.7760}\\
\hline
\end{tabular}}
\label{tab:booktabs}
\end{table}

For the complexity, the total pf parameters and flops are used to evaluate the computational cost and memory consumption of five methods, i.e., VDSR, DnCNN, DRCN, MemNet and ACNet for SISR as described in Table VIII, where the red and blue lines are the best and second results. Although the ACNet refers to more parameters than that of other popular methods, it has less flops. Thus, it is very competitive to state-of-the-arts in SISR.

Execution speed is very important for real applications, such as phones and cameras. Based this reason, we choose five SR methods, such as the VDSR, DRRN, MemNet, CARN-M and ACNet on the recovered LR image of sizes $128 \times 128$, $256 \times 256$ and $512 \times 512$ for $\times 2$  to conduct experiments. As shown in Table IX, it is known that the proposed ACNet is very perferable to other excellent SR methods in running time, where red and blue lines express the first and second SR effect, respectively. That also shows that our proposed ACNet has fast execution speed in SISR.

Perceptual vision is an essential index for interacting between humans and cameras. Motivated by that, we measure the FSIM values between six SR methods, including A+, SelfEx, SRCNN, CARN-M, LESRCNN and ACNet, and the give HR images for $ \times 2$, $\times 3$ and $\times 4$ on the B100, respectively. As shown in Table X, our ACNet has obtained the best results in FSIM than that of other popular SR methods for three scales factors ($ \times 2$, $\times 3$ and $\times 4$). That illustrates that our ACNet has better perceptual vision effect. Additionally, to make perceptual results more vivid, the best and second results are marked by red line and blue line, respectively.

\begin{table}[t!]
\caption{Complexity of five methods in SISR.}
\label{tab:1}
\centering
\scalebox{0.8}[0.8]{
\begin{tabular}{|c|c|c|}
\hline
Methods	&Parameters	&Flops\\
\hline
VDSR\cite{kim2016accurate}	&\textcolor{blue}{665K}	&10.90G\\
\hline
DnCNN\cite{zhang2017beyond}	&\textcolor{red}{556K}	&\textcolor{blue}{9.18G}\\
\hline
DRCN\cite{kim2016deeply}	&1,774K	&29.07G\\
\hline
MemNet\cite{tai2017memnet}	&677K	&11.09G\\
\hline
ACNet (Ours) &1,283K &\textcolor{red}{8.09G}\\
\hline
\end{tabular}}
\label{tab:booktabs}
\end{table}
\begin{table}[t!]
\caption{Running time (seconds) of five methods on the given LR images of sizes $128\times128$, $256\times256$ and $512\times512$ for scale factor of $\times2$.}
\label{tab:1}
\centering
\scalebox{0.74}[0.76]{
\begin{tabular}{|c|c|c|c|c|c|}
\hline
\multicolumn{6}{|c|}{Single Image Super-Resolution} \\
\hline
Size &VDSR\cite{kim2016accurate} &DRRN\cite{tai2017image} &MemNet\cite{tai2017memnet} &CARN-M\cite{ahn2018fast} &ACNet (Ours)\\
\hline
$256 \times 256$ &0.0172	&3.063	&0.8774 &\textcolor{red}{0.0159} &\textcolor{blue}{0.0166}\\
\hline
$512 \times 512$ &0.0575	&8.050	&3.605	&\textcolor{blue}{0.0199}	&\textcolor{red}{0.0195}\\
\hline
$1024 \times 1024$ &0.2126	&25.23	&14.69 &\textcolor{blue}{0.0320}	&\textcolor{red}{0.0315}\\
\hline
\end{tabular}}
\label{tab:booktabs}
\end{table}

\begin{table}[htbp!]
\caption{Average FSIM values of different methods with three scale factors of $\times 2$, $\times 3$ and $\times 4$ on the  B100.}
\label{tab:1}
\centering
\scalebox{0.89}[0.90]{
\begin{tabular}{|c|c|c|c|c|}
\hline
\multirow{1}{*}{Dataset} &
\multirow{1}{*}{Methods} &
$\times 2$ & $\times 3$ & $\times 4$\\
\hline
\multirow{6}{*}{B100} &
A+\cite{timofte2014a+}	&0.9851	&0.9734	&0.9592\\
\cline{2-5} &
SelfEx\cite{huang2015single} &0.9976	&0.9894	&0.9760\\
\cline{2-5} &
SRCNN\cite{dong2014learning} &0.9974	&0.9882	&0.9712\\
\cline{2-5} &
CARN-M\cite{ahn2018fast} &\textcolor{blue}{0.9979}	&0.9898	&0.9765\\
\cline{2-5} &
LESRCNN\cite{tian2020lightweight} &\textcolor{blue}{0.9979}	&\textcolor{blue}{0.9903}	&\textcolor{blue}{0.9774}\\
\cline{2-5} &
ACNet (Ours) &\textcolor{red}{0.9980} &\textcolor{red}{0.9905} &\textcolor{red}{0.9777}\\
\hline
\end{tabular}}
\label{tab:booktabs}
\end{table}

\begin{table}[htbp!]
\caption{Average PSNR/SSIM values of different methods for noise level of 15 with three scale factors of $\times 2$, $\times 3$ and $\times 4$ on the Set5, Set14, B100 and U100.}
\label{tab:1}
\centering
\scalebox{0.89}[0.90]{
\begin{tabular}{|c|c|c|c|c|}
\hline
Noise level &\multicolumn{4}{|c|}{$\sigma = 15$} \\
\cline{1-5} 
\multirow{2}{*}{Dataset} &
\multirow{2}{*}{Methods} &
$\times 2$ & $\times 3$ & $\times 4$\\
\cline{3-5} & &PSNR/SSIM &PSNR/SSIM &PSNR/SSIM\\
\hline
\multirow{4}{*}{Set5} &
DnCNN\cite{zhang2017beyond}	&26.80/0.8080	&25.24/0.7535	&24.59/0.7248\\
\cline{2-5} &
LESRCNN\cite{tian2020lightweight} &\textcolor{blue}{32.67/0.8954}	&\textcolor{blue}{30.36/0.8554}	&\textcolor{red}{28.72}/\textcolor{blue}{0.8175}\\
\cline{2-5} &
ACNet-M (Ours) &\textcolor{red}{32.72/0.8963} &\textcolor{red}{30.39/0.8567} &\textcolor{blue}{28.71}/\textcolor{red}{0.8177}\\
\hline
\multirow{4}{*}{Set14} &
DnCNN\cite{zhang2017beyond}	&25.15/0.7304	&\textcolor{blue}{23.86}/0.6490	&23.09/0.6024\\
\cline{2-5} &
LESRCNN\cite{tian2020lightweight} &\textcolor{blue}{30.22/0.8367}	&\textcolor{red}{27.98}/\textcolor{blue}{0.7619}	&\textcolor{blue}{26.57/0.7049}\\
\cline{2-5} &
ACNet-M (Ours) &\textcolor{red}{30.25/0.8378} &\textcolor{red}{27.98/0.7621} &\textcolor{red}{26.59/0.7056}\\
\hline
\multirow{4}{*}{B100} &
DnCNN\cite{zhang2017beyond}	&25.67/0.6977	&24.58/0.6117	&23.80/0.5708\\
\cline{2-5} &
LESRCNN\cite{tian2020lightweight} &\textcolor{blue}{29.19/0.8041}	&\textcolor{blue}{27.12/0.7148}	&\textcolor{blue}{25.96/0.6553}\\
\cline{2-5} &
ACNet-M (Ours) &\textcolor{red}{29.24/0.8058} &\textcolor{red}{27.15/0.7157} &\textcolor{red}{25.98/0.6563}\\
\hline
\multirow{4}{*}{U100} &
DnCNN\cite{zhang2017beyond}	&23.57/0.6998	&22.19/0.6129	&21.28/0.5567\\
\cline{2-5} &
LESRCNN\cite{tian2020lightweight} &\textcolor{blue}{28.58/0.8561}	&\textcolor{blue}{25.81/0.7711}	&\textcolor{blue}{24.24/0.7029}\\
\cline{2-5} &
ACNet-M (Ours) &\textcolor{red}{28.77/0.8603} &\textcolor{red}{25.92/0.7755} &\textcolor{red}{24.33/0.7078}\\
\hline
\end{tabular}}
\label{tab:booktabs}
\end{table}

\begin{table}[htbp!]
\caption{Average PSNR/SSIM values of different methods for noise level of 25 with three scale factors of $\times 2$, $\times 3$ and $\times 4$ on the Set5, Set14, B100 and U100.}
\label{tab:1}
\centering
\scalebox{0.89}[0.90]{
\begin{tabular}{|c|c|c|c|c|}
\hline
Noise level &\multicolumn{4}{|c|}{$\sigma = 25$} \\
\cline{1-5} 
\multirow{2}{*}{Dataset} &
\multirow{2}{*}{Methods} &
$\times 2$ & $\times 3$ & $\times 4$\\
\cline{3-5} & &PSNR/SSIM &PSNR/SSIM &PSNR/SSIM\\
\hline
\multirow{4}{*}{Set5} &
DnCNN\cite{zhang2017beyond}	&26.14/0.7444	&24.73/0.7064	&24.12/0.6922\\
\cline{2-5} &
LESRCNN\cite{tian2020lightweight} &\textcolor{blue}{30.97/0.8659}	&\textcolor{blue}{28.84/0.8221}	&\textcolor{red}{27.40}/\textcolor{blue}{0.7845}\\
\cline{2-5} &
ACNet-M (Ours) &\textcolor{red}{31.03/0.8672} &\textcolor{red}{28.89/0.8234} &\textcolor{red}{27.40/0.7848}\\
\hline
\multirow{4}{*}{Set14} &
DnCNN\cite{zhang2017beyond}	&24.68/0.6749	&23.47/0.6094	&22.76/0.5740\\
\cline{2-5} &
LESRCNN\cite{tian2020lightweight} &\textcolor{blue}{28.90/0.7941}	&\textcolor{blue}{26.92/0.7217}	&\textcolor{blue}{25.65/0.6695}\\
\cline{2-5} &
ACNet-M (Ours) &\textcolor{red}{28.95/0.7970} &\textcolor{red}{26.93/0.7234} &\textcolor{red}{25.67/0.6710}\\
\hline
\multirow{4}{*}{B100} &
DnCNN\cite{zhang2017beyond}	&25.11/0.6410	&24.13/0.5723	&23.39/0.5415\\
\cline{2-5} &
LESRCNN\cite{tian2020lightweight} &\textcolor{blue}{27.95/0.9528}	&\textcolor{blue}{26.20/0.6718}	&\textcolor{blue}{25.18/0.6203}\\
\cline{2-5} &
ACNet-M (Ours) &\textcolor{red}{28.00/0.7557} &\textcolor{red}{26.24/0.6737} &\textcolor{red}{25.21/0.6221}\\
\hline
\multirow{4}{*}{U100} &
DnCNN\cite{zhang2017beyond}	&23.25/0.6475	&21.93/0.5750	&21.07/0.5296\\
\cline{2-5} &
LESRCNN\cite{tian2020lightweight} &\textcolor{blue}{27.39/0.8201}	&\textcolor{blue}{24.95/0.7348}	&\textcolor{blue}{23.54/0.6698}\\
\cline{2-5} &
ACNet-M (Ours) &\textcolor{red}{27.57/0.8257} &\textcolor{red}{25.06/0.7402} &\textcolor{red}{23.63/0.6751}\\
\hline
\end{tabular}}
\label{tab:booktabs}
\end{table}

\begin{table}[htbp!]
\caption{Average PSNR/SSIM values of different methods for noise level of 35 with three scale factors of $\times 2$, $\times 3$ and $\times 4$ on the Set5, Set14, B100 and U100.}
\label{tab:1}
\centering
\scalebox{0.89}[0.90]{
\begin{tabular}{|c|c|c|c|c|}
\hline
Noise level &\multicolumn{4}{|c|}{$\sigma = 35$} \\
\cline{1-5} 
\multirow{2}{*}{Dataset} &
\multirow{2}{*}{Methods} &
$\times 2$ & $\times 3$ & $\times 4$\\
\cline{3-5} & &PSNR/SSIM &PSNR/SSIM &PSNR/SSIM\\
\hline
\multirow{4}{*}{Set5} &
DnCNN\cite{zhang2017beyond}	&25.32/0.6743	&24.10/0.6515	&23.53/0.6513\\
\cline{2-5} &
LESRCNN\cite{tian2020lightweight} &\textcolor{blue}{29.71/0.8411}	&\textcolor{blue}{27.71/0.7948}	&\textcolor{red}{26.40}/\textcolor{blue}{0.7577}\\
\cline{2-5} &
ACNet-M (Ours) &\textcolor{red}{29.80/0.8433} &\textcolor{red}{27.78/0.7967} &\textcolor{red}{26.41/0.7587}\\
\hline
\multirow{4}{*}{Set14} &
DnCNN\cite{zhang2017beyond}	&24.06/0.6123	&22.96/0.5626	&22.33/0.5390\\
\cline{2-5} &
LESRCNN\cite{tian2020lightweight} &\textcolor{blue}{27.90/0.7589}	&\textcolor{blue}{26.08/0.6899}	&\textcolor{blue}{24.89/0.6419}\\
\cline{2-5} &
ACNet-M (Ours) &\textcolor{red}{27.96/0.7635} &\textcolor{red}{26.11/0.6927} &\textcolor{red}{24.93/0.6441}\\
\hline
\multirow{4}{*}{B100} &
DnCNN\cite{zhang2017beyond}	&24.40/0.5779	&23.55/0.5262	&22.86/0.5056\\
\cline{2-5} &
LESRCNN\cite{tian2020lightweight} &\textcolor{blue}{27.05/0.7136}	&\textcolor{blue}{25.50/0.6402}	&\textcolor{blue}{24.57/0.5950}\\
\cline{2-5} &
ACNet-M (Ours) &\textcolor{red}{27.10/0.7175} &\textcolor{red}{25.55/0.6426} &\textcolor{red}{24.61/0.5970}\\
\hline
\multirow{4}{*}{U100} &
DnCNN\cite{zhang2017beyond}	&22.80/0.5906	&21.56/0.5312	&20.76/0.4968\\
\cline{2-5} &
LESRCNN\cite{tian2020lightweight} &\textcolor{blue}{26.44}/\textcolor{red}{0.7980}	&\textcolor{blue}{24.24/0.7040}	&\textcolor{blue}{22.96/0.6425}\\
\cline{2-5} &
ACNet-M (Ours) &\textcolor{red}{26.63}/\textcolor{blue}{0.7949} &\textcolor{red}{24.36/0.7102} &\textcolor{red}{23.05/0.6482}\\
\hline
\end{tabular}}
\label{tab:booktabs}
\end{table}

\begin{table}[htbp!]
\caption{Average PSNR/SSIM values of different methods for noise level of 50 with three scale factors of $\times 2$, $\times 3$ and $\times 4$ on the Set5, Set14, B100 and U100.}
\label{tab:1}
\centering
\scalebox{0.89}[0.90]{
\begin{tabular}{|c|c|c|c|c|}
\hline
Noise level &\multicolumn{4}{|c|}{$\sigma = 50$} \\
\cline{1-5} 
\multirow{2}{*}{Dataset} &
\multirow{2}{*}{Methods} &
$\times 2$ & $\times 3$ & $\times 4$\\
\cline{3-5} & &PSNR/SSIM &PSNR/SSIM &PSNR/SSIM\\
\hline
\multirow{4}{*}{Set5} &
DnCNN\cite{zhang2017beyond}	&24.02/0.5751	&23.05/0.5686	&22.56/0.5849\\
\cline{2-5} &
LESRCNN\cite{tian2020lightweight} &\textcolor{blue}{28.29/0.8088}	&\textcolor{blue}{26.41/0.7602}	&\textcolor{blue}{25.23/0.7245}\\
\cline{2-5} &
ACNet-M (Ours) &\textcolor{red}{28.37/0.8114} &\textcolor{red}{26.51/0.7628} &\textcolor{red}{25.25/0.7254}\\
\hline
\multirow{4}{*}{Set14} &
DnCNN\cite{zhang2017beyond}	&23.01/0.5224	&22.09/0.4912	&21.57/0.4827\\
\cline{2-5} &
LESRCNN\cite{tian2020lightweight} &\textcolor{blue}{26.74/0.7166}	&\textcolor{blue}{25.09/0.6535}	&\textcolor{blue}{24.01/0.6111}\\
\cline{2-5} &
ACNet-M (Ours) &\textcolor{red}{26.81/0.7220} &\textcolor{red}{25.14/0.6566} &\textcolor{red}{24.03/0.6122}\\
\hline
\multirow{4}{*}{B100} &
DnCNN\cite{zhang2017beyond}	&23.20/0.4884	&22.54/0.4568	&21.93/0.4486\\
\cline{2-5} &
LESRCNN\cite{tian2020lightweight} &\textcolor{blue}{26.06/0.6697}	&\textcolor{blue}{24.68/0.6056}	&\textcolor{blue}{23.85/0.5675}\\
\cline{2-5} &
ACNet-M (Ours) &\textcolor{red}{26.11/0.6742} &\textcolor{red}{24.75/0.6082} &\textcolor{red}{23.89/0.5693}\\
\hline
\multirow{4}{*}{U100} &
DnCNN\cite{zhang2017beyond}	&21.95/0.5114	&20.87/0.4663	&20.17/0.4453\\
\cline{2-5} &
LESRCNN\cite{tian2020lightweight} &\textcolor{blue}{25.32/0.7459}	&\textcolor{blue}{23.38/0.6653}	&\textcolor{blue}{22.24/0.6089}\\
\cline{2-5} &
ACNet-M (Ours) &\textcolor{red}{25.50/0.7539} &\textcolor{red}{23.50/0.6720} &\textcolor{red}{22.33/0.6149}\\
\hline
\end{tabular}}
\label{tab:booktabs}
\end{table}

According to these presentations, we can see that the proposed ACNet obtains a good SR performance in quantitative analysis.




Qualitative analysis: To comprehensively test the visual quality of our ACNet, we construct the predicted SR images, Y channel images of obtained SR images, errors images and edge images in flat areas, texture areas and detailed areas to observe  visual effect of different methods for SISR, respectively, where edge images are obtained by Canny method \cite{bao2005canny}. 
Specifically, the observed area is one amplified area of the predicted SR image. That is clearer besides in the error image, the corresponding method can obtain better performance in SISR. In the error images, the observed areas are more blur, the corresponding SR methods achieve more excellent results. Specifically, Figs. 3-6 denote the visual effect of different SR methods on predicted SR images, Y channel images, error images and edge images for recovering flat areas, respectively. Moreover, Figs. 7-10 express the vivid effect of different SR methods on predicted SR images, Y channel images, error images and edge images for recovering texture areas, respectively. Besides, Figs. 11-14 show the high-quality images  of different SR methods on predicted SR images, Y channel images, error images and edge images for recovering detailed areas, respectively. Figs. 3-14 are shown at https://github.com/hellloxiaotian/ACNet/blob/master/figs.pdf. By these figures, we can see that the proposed ACNet has strong robustness in gaining clearer images.

\subsection{Extensions}
It is known that captured images simultaneously suffer from multiple factors, such as unknown noise and unknown scale factor by camera and shooting environment. However, there is little effort to tackle this problem. Hence, we propose ACNet to tackle a multiple-degradation task (i.e., blind SR with blind noise) as well as ACNet-M via $y = x \downarrow {_s} + \upsilon$, where $y$ and $x$ represent the LR and HR images, respectively,  $s$ is scale factor, and $\upsilon$ is additive white Gaussian noise with a noise level of $\sigma$ ranging from 0 to 55 during training. Moreover, the other training parameters are the same as both ACNet and ACNet-B in Section IV. C, training and test datasets are the same as both ACNet and ACNet-B in Sections IV. A and B. Additionally, we compare our method with DnCNN  and LESRCNN by conducting SR experiments with noise levels of 15, 25, 35 and 50 for $\times 2$, $\times 3$ and $\times 4$ on Set5, Set14, B100, and U100. Tables XI-XIV show that ACNet-M is competitive with DnCNN and LESRCNN for blind SISR with blind noise. For instance, our ACNet-M outperforms LESRCNN by 0.19dB in PSNR and 0.0038 in SSIM on U100 for $\times 2$ when $\upsilon=15$ in Table XI. Although ACNet only outperforms CARN-M  by a small margin on Set 5 for $\times 3$, ACNet achieves good performance in all aspects of computational cost, run-time complexity, perceptual analysis, and visual quality. Additionally, it is versatile that its model can handle various SR tasks such as blind SISR with blind noise.

\section{Conclusion}
In this paper, we proposed an asymmetric CNN (ACNet) by an AB, a MEB and a HFFEB. The AB utilizes one-dimensional asymmetric convolutions to intensify the square convolution kernels in horizontal and vertical directions to promote the effects of salient features for single image superresolution. The MEB fuses all hierarchical low-frequency features via the residual learning technique to resolve the long-term dependency problem and transform obtained low-frequency features into high-frequency features. The HFFEB exploits low- and high- frequency features to obtain more robust super-resolution features. Also, to prevent excessive feature enhancement, the FFEB uses extra two-layer convolutions to learn more accurate SR features, which can reduce the difference between the predicted SR image and the given HR image to improve the stability of training processing. Additionally, it takes charge of reconstructing a HR image. Further, experimental results show that the ACNet performs well against state-of-the-art super-resolution methods in terms of both quantitative and qualitative evaluations.
\ifCLASSOPTIONcaptionsoff
  \newpage
\fi
%

\bibliography{chunweitian}

\begin{thebibliography}{1}
\bibitem{liang2019novel}
X.~Liang, D.~Zhang, G.~Lu, Z.~Guo, and N.~Luo, ``A novel multicamera system for
  high-speed touchless palm recognition,'' \emph{IEEE Trans. Syst.,
  Man, Cybern.: Syst.}, 2019.

\bibitem{shi2013cardiac}
W.~Shi, J.~Caballero, C.~Ledig, X.~Zhuang, W.~Bai, K.~Bhatia, A.~M. S.~M.
  de~Marvao, T.~Dawes, D.~O’Regan, and D.~Rueckert, ``Cardiac image
  super-resolution with global correspondence using multi-atlas patchmatch,''
  in \emph{Proc. Int. Conf. Medical Image Comput.
  Comput.-Assisted Intervention}.\hskip 1em plus 0.5em minus 0.4em\relax
  Springer, 2013, pp. 9--16.

\bibitem{du2019improved}
B.~Du, Q.~Wei, and R.~Liu, ``An improved quantum-behaved particle swarm
  optimization for endmember extraction,'' \emph{IEEE Trans.
  Geosci. Remote Sens.}, vol.~57, no.~8, pp. 6003--6017, 2019.

\bibitem{irani1991improving}
M.~Irani and S.~Peleg, ``Improving resolution by image registration,''
  \emph{CVGIP: Graphical Models Image Process.}, vol.~53, no.~3, pp.
  231--239, 1991.

\bibitem{sun2008image}
J.~Sun, Z.~Xu, and H.-Y. Shum, ``Image super-resolution using gradient profile
  prior,'' in \emph{Proc. IEEE/CVF
  Conf. Comput. Vis. Pattern Recognit.}.\hskip 1em plus 0.5em minus 0.4em\relax IEEE, 2008, pp. 1--8.

\bibitem{xiong2010robust}
Z.~Xiong, X.~Sun, and F.~Wu, ``Robust web image/video super-resolution,''
  \emph{IEEE Trans. Image Process.}, vol.~19, no.~8, pp. 2017--2028,
  2010.

\bibitem{chang2004super}
H.~Chang, D.-Y. Yeung, and Y.~Xiong, ``Super-resolution through neighbor
  embedding,'' in \emph{Proc. IEEE/CVF
  Conf. Comput. Vis. Pattern Recognit.},
  vol.~1.\hskip 1em plus 0.5em minus 0.4em\relax IEEE, 2004, pp. I--I.

\bibitem{gao2014novel}
G.~Gao and J.~Yang, ``A novel sparse representation based framework for face
  image super-resolution,'' \emph{Neurocomputing}, vol. 134, pp. 92--99, 2014.


\bibitem{alvarez2018image}
V.~Alvarez, V.~Ponomaryov and S.~Sadovnychiy ``Image super-resolution via wavelet feature extraction and sparse representation,'' \emph{Radioengineering}, vol. 27, pp. 602-609, 2018.

\bibitem{alvarez2018imageb}
V.~Alvarez, V.~Ponomaryov, S.~Sadovnychiy and R. Reyes``Image super-resolution via two coupled dictionaries and sparse representation,'' \emph{Multimedia Tools Appl.}, vol. 77, pp. 13487-13511, 2018.

\bibitem{yang2016single}
W.~Yang, T.~Yuan, W.~Wang, F.~Zhou, and Q.~Liao, ``Single-image
  super-resolution by subdictionary coding and kernel regression,'' \emph{IEEE
  Trans. Syst., Man, Cybern.: Syst.}, vol.~47, no.~9, pp.
  2478--2488, 2016.

\bibitem{schulter2015fast}
S.~Schulter, C.~Leistner, and H.~Bischof, ``Fast and accurate image upscaling
  with super-resolution forests,'' in \emph{Proc. IEEE/CVF
  Conf. Comput. Vis. Pattern Recognit.}, 2015, pp. 3791--3799.

\bibitem{tian2020image}
C.~Tian, Y.~Xu, and W.~Zuo, ``Image denoising using deep cnn with batch
  renormalization,'' \emph{Neural Netw.}, vol. 121, pp. 461--473, 2020.

\bibitem{dong2014learning}
C.~Dong, C.~C. Loy, K.~He, and X.~Tang, ``Learning a deep convolutional network
  for image super-resolution,'' in \emph{Proc. European Conf. Comput.
  Vis.}.\hskip 1em plus 0.5em minus 0.4em\relax Springer, 2014, pp. 184--199.

\bibitem{kim2016accurate}
J.~Kim, J.~Kwon~Lee, and K.~Mu~Lee, ``Accurate image super-resolution using
  very deep convolutional networks,'' in \emph{Proc. IEEE/CVF
  Conf. Comput. Vis. Pattern Recognit.}, 2016, pp. 1646--1654.

\bibitem{kim2016deeply}
J.~Kim, J.~Kwon~Lee, and K.~Mu~Lee, ``Deeply-recursive convolutional network for image super-resolution,''
  in \emph{Proc. IEEE/CVF
  Conf. Comput. Vis. Pattern Recognit.}, 2016, pp. 1637--1645.

\bibitem{tai2017image}
Y.~Tai, J.~Yang, and X.~Liu, ``Image super-resolution via deep recursive
  residual network,'' in \emph{Proc. IEEE/CVF
  Conf. Comput. Vis. Pattern Recognit.}, 2017, pp. 3147--3155.

\bibitem{mao2016image}
X.~Mao, C.~Shen, and Y.-B. Yang, ``Image restoration using very deep
  convolutional encoder-decoder networks with symmetric skip connections,'' in
  \emph{Proc. Adv. Neural Inf. Process.g Syst.}, 2016, pp.
  2802--2810.
  
\bibitem{tian2020lightweight}
C.~Tian, R.~Zhu, Z.~Wu, Y.~Xu, W.~Zuo, C.~Chen, and C.~Lin, ``Lightweight image super-Resolution with enhanced CNN,'' \emph{Knowl.-Based Syst.}, 2020.

\bibitem{tian2019deep}
C.~Tian, L.~Fei, W.~Zheng, Y.~Xu, W.~Zuo, and C.-W. Lin, ``Deep learning on
  image denoising: An overview,'' \emph{arXiv preprint arXiv:1912.13171}, 2019.

\bibitem{fan2018compressed}
X.~Fan, Y.~Yang, C.~Deng, J.~Xu, and X.~Gao, ``Compressed multi-scale feature
  fusion network for single image super-resolution,'' \emph{Signal Process.},
  vol. 146, pp. 50--60, 2018.

\bibitem{zhang2018dcsr}
Z.~Zhang, X.~Wang, and C.~Jung, ``Dcsr: Dilated convolutions for single image
  super-resolution,'' \emph{IEEE Trans. Image Process.}, vol.~28,
  no.~4, pp. 1625--1635, 2018.

\bibitem{zhang2017learning}
K.~Zhang, W.~Zuo, S.~Gu, and L.~Zhang, ``Learning deep cnn denoiser prior for
  image restoration,'' in \emph{Proc. IEEE/CVF
  Conf. Comput. Vis. Pattern Recognit.}, 2017, pp. 3929--3938.

\bibitem{liu2018multi}
P.~Liu, H.~Zhang, K.~Zhang, L.~Lin, and W.~Zuo, ``Multi-level wavelet-cnn for
  image restoration,'' in \emph{Proc. IEEE/CVF
  Conf. Comput. Vis. Pattern Recognit. Workshops}, 2018, pp. 773--782.

\bibitem{dong2016accelerating}
C.~Dong, C.~C. Loy, and X.~Tang, ``Accelerating the super-resolution
  convolutional neural network,'' in \emph{European Conference on Computer
  Vision}.\hskip 1em plus 0.5em minus 0.4em\relax Springer, 2016, pp. 391--407.

\bibitem{ahn2018fast}
N.~Ahn, B.~Kang, and K.-A. Sohn, ``Fast, accurate, and lightweight
  super-resolution with cascading residual network,'' in \emph{Proc. European Conf. Comput. Vis.}, 2018, pp. 252--268.

\bibitem{hui2019lightweight}
Z.~Hui, X.~Gao, Y.~Yang, and X.~Wang, ``Lightweight image super-resolution with
  information multi-distillation network,'' in \emph{Proc.
  ACM Int. Conf. Multimedia}, 2019, pp. 2024--2032.

\bibitem{zhang2018gated}
X.~Zhang, H.~Dong, Z.~Hu, W.-S. Lai, F.~Wang, and M.-H. Yang, ``Gated fusion
  network for joint image deblurring and super-resolution,'' \emph{arXiv
  preprint arXiv:1807.10806}, 2018.

\bibitem{tian2020attention}
C.~Tian, Y.~Xu, Z.~Li, W.~Zuo, L.~Fei, and H.~Liu, ``Attention-guided cnn for
  image denoising,'' \emph{Neural Netw.}, vol. 124, pp. 177--129, 2020.
  
\bibitem{du2018robust}
B.~Du, T.~Xinyao, Z.~Wang, L.~Zhang, and D.~Tao, ``Robust graph-based
  semisupervised learning for noisy labeled data via maximum correntropy
  criterion,'' \emph{IEEE Trans. Cybern.}, vol.~49, no.~4, pp.
  1440--1453, 2018.
  
 \bibitem{wang2020weighted}
C.~Wang, Z.~Yan, W.~Pedrycz, M.~Zhou, and Z.~Li, ``A weighted fidelity and regularization-based method for mixed or unknown noise removal from images on graphs,'' \emph{IEEE Trans. Image Process.}, vol.~29, pp.
  5229--5243, 2020.
  
 \bibitem{ren2019progressive}
D.~Ren, W.~Zuo, Q. Hu, P. Zhu, and Q. Hu, ``Progressive image deraining networks: a better and simpler baseline,'' in \emph{Proc. IEEE/CVF
  Conf. Comput. Vis. Pattern Recognit.}, 2019, pp. 3937--3946.

 \bibitem{ren2020single}
W.~Ren, J.~Pan, H.~Zhang, X.~Cao, and M.~Yang, `Single image dehazing via multi-scale convolutional neural networks with holistic edges,'' \emph{Int. J. Comput. Vis.}, vol.~128, no.~1, pp.
  240--259, 2020.  
  
\bibitem{zhang2019deep}
K.~Zhang, W.~Zuo, and L.~Zhang, ``Deep plug-and-play super-resolution for
  arbitrary blur kernels,'' in \emph{Proc. IEEE/CVF
  Conf. Comput. Vis. Pattern Recognit.}, 2019, pp. 1671--1681.
  
\bibitem{tian2020coarse}
C.~Tian, Y.~Xu, Z.~Wang, W.~Zuo, B.~Zhang, L.~Fei and C.~Lin, ``Coarse-to-fine CNN for image super-resolution,'' \emph{IEEE Trans. Multimedia}, 2020.
 
\bibitem{li2019fast}
S.~Li, F.~He, B.~Du, L.~Zhang, Y.~Xu, and D.~Tao, ``Fast spatio-temporal
  residual network for video super-resolution,'' in \emph{Proc. IEEE/CVF
  Conf. Comput. Vis. Pattern Recognit.}, 2019, pp.
  10\,522--10\,531.

\bibitem{ding2019acnet}
X.~Ding, Y.~Guo, G.~Ding, and J.~Han, ``Acnet: Strengthening the kernel
  skeletons for powerful cnn via asymmetric convolution blocks,'' in
  \emph{Proc. IEEE/CVF
  Conf. Comput. Vis.},
  2019, pp. 1911--1920.

\bibitem{jin2014flattened}
J.~Jin, A.~Dundar, and E.~Culurciello, ``Flattened convolutional neural
  networks for feedforward acceleration,'' \emph{arXiv preprint
  arXiv:1412.5474}, 2014.

\bibitem{szegedy2016rethinking}
C.~Szegedy, V.~Vanhoucke, S.~Ioffe, J.~Shlens, and Z.~Wojna, ``Rethinking the
  inception architecture for computer vision,'' in \emph{Proc. IEEE/CVF
  Conf. Comput. Vis. Pattern Recognit.}, 2016, pp.
  2818--2826.

\bibitem{lo2019efficient}
S.-Y. Lo, H.-M. Hang, S.-W. Chan, and J.-J. Lin, ``Efficient dense modules of
  asymmetric convolution for real-time semantic segmentation,'' in
  \emph{Proc. ACM Multimedia Asia}, 2019, pp. 1--6.

\bibitem{lai2017deep}
W.-S. Lai, J.-B. Huang, N.~Ahuja, and M.-H. Yang, ``Deep {}Laplacian pyramid
  networks for fast and accurate super-resolution,'' in \emph{Proc. IEEE/CVF
  Conf. Comput. Vis. Pattern Recognit.}, 2017, pp.
  624--632.

\bibitem{hui2018fast}
Z.~Hui, X.~Wang, and X.~Gao, ``Fast and accurate single image super-resolution
  via information distillation network,'' in \emph{Proc. IEEE/CVF
  Conf. Comput. Vis. Pattern Recognit.}, 2018, pp. 723--731.


\bibitem{xie2018fast}
C.~Xie, W.~Zeng, and X.~Lu, ``Fast single-image super-resolution via deep
  network with component learning,'' \emph{IEEE Trans. Circuits
  Syst. Video Technol.}, vol.~29, no.~12, pp. 3473--3486, 2018.

\bibitem{shi2016real}
W.~Shi, J.~Caballero, F.~Huszar, J.~Totz, A.~Aitken, B.~Bishop, D.~Rueckert, and Z.~Wang,``Real-time single image and video super-resolution using an efficient sub-pixel convolutional neural network,'' in \emph{Proc. IEEE/CVF
  Conf. Comput. Vis. Pattern Recognit.}, 2016, pp. 1874--1883.

\bibitem{yang2019deep}
W.~Yang, X.~Zhang, Y.~Tian, W.~Wang, J.-H. Xue, and Q.~Liao, ``Deep learning
  for single image super-resolution: A brief review,'' \emph{IEEE Trans. Multimedia}, vol.~21, no.~12, pp. 3106--3121, 2019.
  
\bibitem{douillard1995iterative}
C.~Douillard, M.~J{\'e}z{\'e}quel, C.~Berrou, D.~Electronique, A.~Picart,
  P.~Didier, and A.~Glavieux, ``Iterative correction of intersymbol
  interference: Turbo-equalization,'' \emph{European Trans.
  Telecommun.}, vol.~6, no.~5, pp. 507--511, 1995.
  
\bibitem{krizhevsky2012imagenet}
A.~Krizhevsky, I.~Sutskever, and G.~E. Hinton, ``Imagenet classification with
  deep convolutional neural networks,'' in \emph{Proc. Adv. Neural Inf.
  Process. Syst.}, 2012, pp. 1097--1105.

\bibitem{tai2017memnet}
Y.~Tai, J.~Yang, X.~Liu, and C.~Xu, ``Memnet: A persistent memory network for
  image restoration,'' in \emph{Proc. IEEE/CVF Int.
  Conf. Comput. Vis.}, 2017, pp. 4539--4547.

\bibitem{lim2017enhanced}
B.~Lim, S.~Son, H.~Kim, S.~Nah, and K.~Mu~Lee, ``Enhanced deep residual
  networks for single image super-resolution,'' in \emph{Proc. IEEE/CVF Conf. Comput. Vis. Pattern
  Recognit. Workshops}, 2017,
  pp. 136--144.

\bibitem{timofte2017ntire}
R.~Timofte, E.~Agustsson, L.~Van~Gool, M.-H. Yang, and L.~Zhang, ``Ntire 2017
  challenge on single image super-resolution: Methods and results,'' in
  \emph{Proc. IEEE/CVF Conf. Comput. Vis. Pattern
  Recognit. Workshops}, 2017, pp. 114--125.

\bibitem{tian2020designing}
C.~Tian, Y.~Xu, W.~Zuo, B.~Du, C.-W. Lin and D. ~Zhang, ``Designing and Training of A Dual CNN for Image Denoising,'' \emph{arXiv preprint arXiv:2007.03951}, 2020.

\bibitem{yuan2020visual}
D.~Yuan, X.~Li, Z.~He, Q.~Liu, and S.~Lu, ``Visual object tracking with
  adaptive structural convolutional network,'' \emph{Knowl.-Based Syst.},
  p. 105554, 2020.

\bibitem{bevilacqua2012low}
M.~Bevilacqua, A.~Roumy, C.~Guillemot, and M.~L. Alberi-Morel, ``Low-complexity
  single-image super-resolution based on nonnegative neighbor embedding,''
  in
  \emph{Proc. British Machi. Vis. Conf.}, 2012.
  
\bibitem{yang2010image}
J.~Yang, J.~Wright, T.~S. Huang, and Y.~Ma, ``Image super-resolution via sparse
  representation,'' \emph{IEEE Trans. Image Process.}, vol.~19,
  no.~11, pp. 2861--2873, 2010.

\bibitem{martin2001database}
D.~Martin, C.~Fowlkes, D.~Tal, and J.~Malik, ``A database of human segmented
  natural images and its application to evaluating segmentation algorithms and
  measuring ecological statistics,'' in \emph{Proc. IEEE/CVF
  Int. Conf. Comput. Vis.}, vol.~2.\hskip 1em
  plus 0.5em minus 0.4em\relax IEEE, 2001, pp. 416--423.

\bibitem{huang2015single}
J.-B. Huang, A.~Singh, and N.~Ahuja, ``Single image super-resolution from
  transformed self-exemplars,'' in \emph{Proc. IEEE/CVF Conf.
  Comput. Vis. Pattern Recognit.}, 2015, pp. 5197--5206.

\bibitem{kingma2014adam}
D.~P. Kingma and J.~Ba, ``Adam: A method for stochastic optimization,''
  \emph{arXiv preprint arXiv:1412.6980}, 2014.

\bibitem{zhang2017beyond}
K.~Zhang, W.~Zuo, Y.~Chen, D.~Meng, and L.~Zhang, ``Beyond a gaussian denoiser:
  Residual learning of deep {CNN} for image denoising,'' \emph{IEEE Trans. Image Process.}, vol.~26, no.~7, pp. 3142--3155, 2017.

\bibitem{zhang2018residual}
Y.~Zhang, Y.~Tian, Y.~Kong, B.~Zhong, and Y.~Fu, ``Residual dense network for
  image super-resolution,'' in \emph{Proc. IEEE/CVF Conf.
  Comput. Vis. Pattern Recognit.}, 2018, pp. 2472--2481.

\bibitem{hore2010image}
A.~Hore and D.~Ziou, ``Image quality metrics: {PSNR} vs. {SSIM},'' in \emph{2010
  Proc. Int. Conf. Pattern Recognit.}.\hskip 1em plus 0.5em
  minus 0.4em\relax IEEE, 2010, pp. 2366--2369.

\bibitem{he2016deep}
K.~He, X.~Zhang, S.~Ren, and J.~Sun, ``Deep residual learning for image
  recognition,'' in \emph{Proc. IEEE/CVF Conf. Comput. Vis. Pattern Recognit.}, 2016, pp. 770--778.

\bibitem{ahn2018image}
N.~Ahn, B.~Kang and K.~Sohn, ``Image super-resolution via progressive cascading residual network,'' in \emph{Proceedings of the IEEE Conference on Computer Vision and Pattern Recognition}, 2018, pp. 791--799.
  
  
\bibitem{zhang2011fsim}
L.~Zhang, L.~Zhang, X.~Mou and D.~Zhang, ``{FSIM}: A feature similarity index for image quality assessment,'' \emph{IEEE Trans. Image Process.}, vol.~20, no.~8, pp.
  2378--2386, 2011.


\bibitem{timofte2014a+}
R.~Timofte, V.~De~Smet, and L.~Van~Gool, ``A+: Adjusted anchored neighborhood
  regression for fast super-resolution,'' in \emph{Proc. Asian Conf. Comput.
  Vis.}.\hskip 1em plus 0.5em minus 0.4em\relax Springer, 2014, pp. 111--126.

\bibitem{dai2015jointly}
D.~Dai, R.~Timofte, and L.~Van~Gool, ``Jointly optimized regressors for image
  super-resolution,'' in \emph{Comput. Graphics Forum}, vol.~34, no.~2.\hskip
  1em plus 0.5em minus 0.4em\relax Wiley Online Library, 2015, pp. 95--104.

\bibitem{wang2015deep}
Z.~Wang, D.~Liu, J.~Yang, W.~Han, and T.~Huang, ``Deep networks for image
  super-resolution with sparse prior,'' in \emph{Proc. IEEE/CVF
 Int. Conf. Comput. Vis.}, 2015, pp. 370--378.

\bibitem{chen2016trainable}
Y.~Chen and T.~Pock, ``Trainable nonlinear reaction diffusion: A flexible
  framework for fast and effective image restoration,'' \emph{IEEE Trans. Pattern Anal. Mach. Intell.}, vol.~39, no.~6, pp.
  1256--1272, 2016.

\bibitem{lu2018fast}
Z.~Lu, Z.~Yu, P.~Yali, L.~Shigang, W.~Xiaojun, L.~Gang, and R.~Yuan, ``Fast
  single image super-resolution via dilated residual networks,'' \emph{IEEE
  Access}, vol.~7, pp. 109\,729--109\,738, 2018.

\bibitem{shi2017structure}
Y.~Shi, K.~Wang, C.~Chen, L.~Xu, and L.~Lin, ``Structure-preserving image
  super-resolution via contextualized multitask learning,'' \emph{IEEE
  Trans. Multimedia}, vol.~19, no.~12, pp. 2804--2815, 2017.

\bibitem{ren2017image}
H.~Ren, M.~El-Khamy, and J.~Lee, ``Image super resolution based on fusing
  multiple convolution neural networks,'' in \emph{Proc. IEEE/CVF
  Conf. Comput. Vis. Pattern Recognit. Workshops}, 2017, pp.
  54--61.

\bibitem{bae2017beyond}
W.~Bae, J.~Yoo, and J.~Chul~Ye, ``Beyond deep residual learning for image
  restoration: Persistent homology-guided manifold simplification,'' in
  \emph{Proc. IEEE/CVF
  Conf. Comput. Vis. Pattern Recognit. Workshops}, 2017, pp. 145--153.

\bibitem{xu2018self}
J.~Xu, M.~Li, J.~Fan, X.~Zhao, and Z.~Chang, ``Self-learning super-resolution
  using convolutional principal component analysis and random matching,''
  \emph{IEEE Trans. Multimedia}, vol.~21, no.~5, pp. 1108--1121, 2018.

\bibitem{yang2018drfn}
X.~Yang, H.~Mei, J.~Zhang, K.~Xu, B.~Yin, Q.~Zhang, and X.~Wei, ``Drfn: Deep
  recurrent fusion network for single-image super-resolution with large
  factors,'' \emph{IEEE Trans. Multimedia}, vol.~21, no.~2, pp.
  328--337, 2018.

\bibitem{cruz2017single}
C.~Cruz, R.~Mehta, V.~Katkovnik, and K.~O. Egiazarian, ``Single image
  super-resolution based on wiener filter in similarity domain,'' \emph{IEEE
  Trans. Image Process.}, vol.~27, no.~3, pp. 1376--1389, 2017.

\bibitem{cao2019new}
F.~Cao and B.~Chen, ``New architecture of deep recursive convolution networks for super-resolution,'' \emph{Knowledge-Based Syst.}, vol. 178, pp. 98--110, 2019.


\bibitem{fan2017balanced}
Y.~Fan, H.~Shi, J.~Yu, D.~Liu, W.~Han, H.~Yu, Z.~Wang, X.~Wang, and T.~S.
  Huang, ``Balanced two-stage residual networks for image super-resolution,''
  in \emph{Proc. IEEE/CVF
  Conf. Comput. Vis. Pattern Recognit. Workshops}, 2017, pp. 161--168.
  
\bibitem{bao2005canny}
P.~Bao, L.~Zhang and X.~Wu, ``Canny edge detection enhancement by scale multiplication,''
  in \emph{IEEE Trans. Pattern Anal. Mach. Intell.}, vol.~27, no.~9, pp. 1485--1490,2005.

\end{thebibliography}
\begin{IEEEbiography}[{\includegraphics[width=0.9in,height=1.2in]{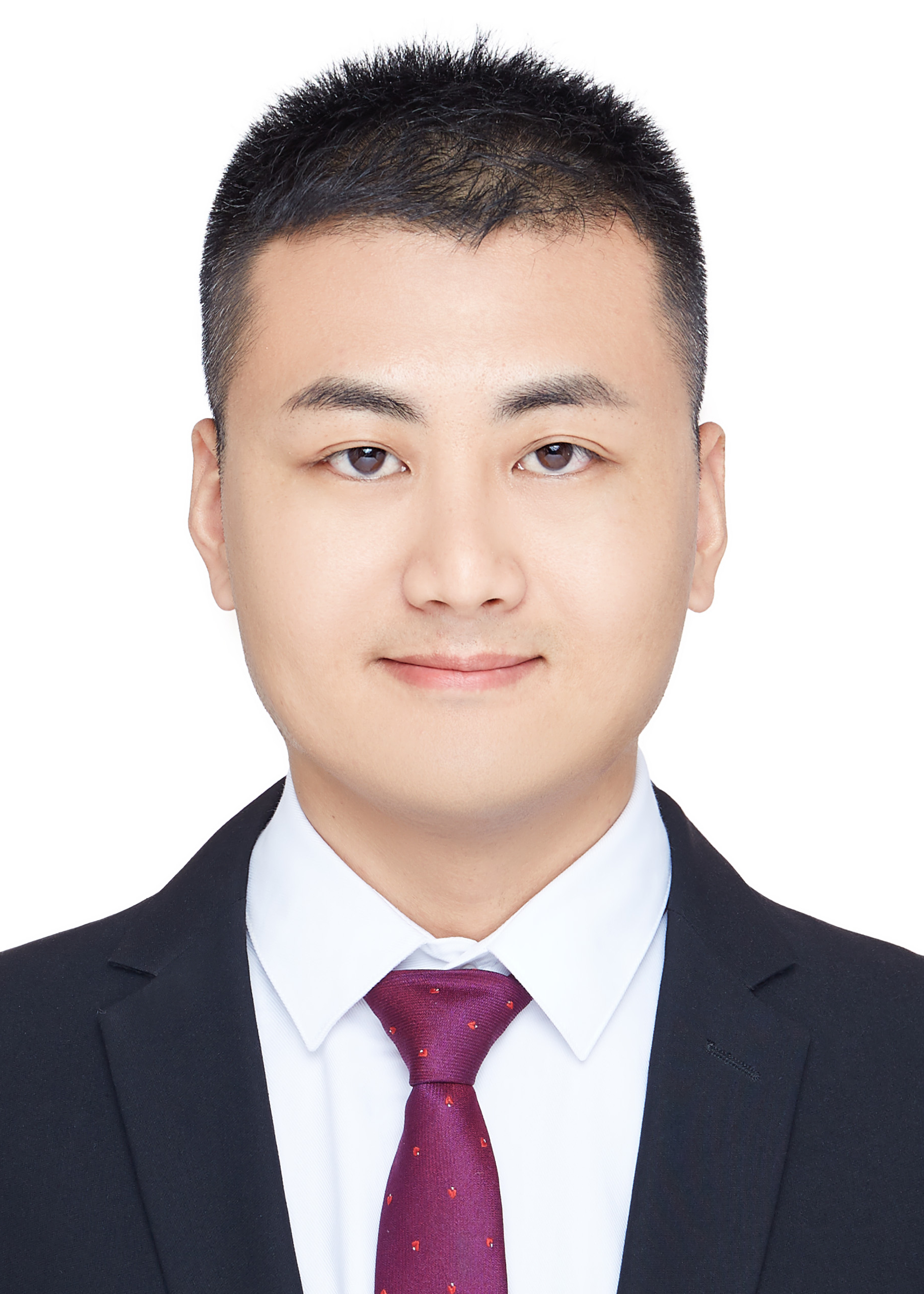}}]{Chunwei Tian} (Member, IEEE)
received his Ph.D degree
in the School of Computer Science and Technology
at Harbin Institute of Technology, in Jan, 2021. His research interests include image restoration and deep learning. He has published over 30 papers in academic journals and conferences, including IEEE TNNLS, IEEE TMM, NN, Information Sciences, KBS, PRL, ICASSP, ICPR, ACPR and IJCB. He is an associate editor of the Journal of Electrical and Electronic Engineering, a PC of the IEEE DASC 2020, a PC Assistant of IJCAI 2019, a reviewer of some journals and conferences, such as the IEEE TIP, the IEEE TII, the IEEE TMM, the IEEE TSMC, the NN, the CVIU, the Neurocomputing, the Visual Computer, the PRL and the SPL, etc.
\end{IEEEbiography}

\begin{IEEEbiography}[{\includegraphics[width=1.0in,height=1.2in]{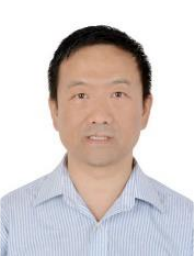}}]{Yong Xu} (Senior Member,  IEEE) received the
Ph.D. degree in Pattern Recognition and Intelligence system at NUST (China) in 2005. Now he works at Harbin Institute of Technology. His current interests include pattern recognition, deep learning, image processing. He has published over 100 papers in known academic journals and conferences. He has served as an Co-Editors-in-Chief of the International Journal
of Image and Graphics, an Associate Editor of the CAAI Transactions on Intelligence Technology.
\end{IEEEbiography}

\begin{IEEEbiography}[{\includegraphics[width=0.9in,height=1.2in]{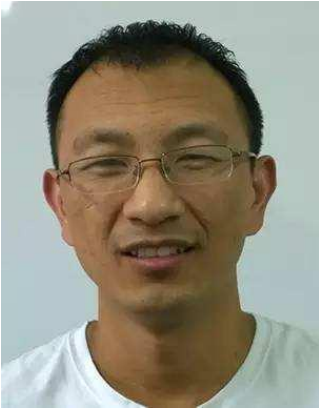}}]{Wangmeng Zuo} (Senior Member,  IEEE)
received the Ph.D. degree in computer application technology from the Harbin Institute of Technology, Harbin, China, in 2007. He is currently a Professor in the School of Computer
Science and Technology, Harbin Institute of Technology. His current research interests include image enhancement and restoration, object detection, visual tracking, and image classification.
He has published over 100 papers in toptier academic journals and conferences. He has served as a Tutorial Organizer in ECCV 2016, an Associate Editor of the IET Biometrics and Journal of Electronic Imaging, and the Guest Editor of Neurocomputing, Pattern Recognition, IEEE Transactions on Circuits and Systems for Video Technology, and IEEE Transactions on Neural Networks and Learning Systems.
\end{IEEEbiography}

\begin{IEEEbiography}[{\includegraphics[width=0.9in,height=1.2in]{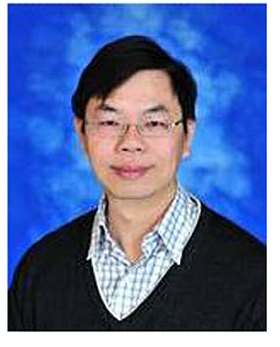}}]{Chia-Wen Lin}
(Fellow,  IEEE) received his Ph.D. degree in electrical engineering from
National Tsing Hua University (NTHU), Hsinchu,
Taiwan, in 2000. He is currently Professor with the Department of
Electrical Engineering and the Institute of Communications Engineering, NTHU. He is also Deputy
Director of the AI Research Center of NTHU. He was with the Department of Computer Science and
Information Engineering, National Chung Cheng
University, Taiwan, during 2000${\rm{ - }}$2007. Prior to joining academia, he worked for the Information and Communications Research Laboratories, Industrial Technology Research Institute, Hsinchu, Taiwan,
during 1992${\rm{ - }}$2000. His research interests include image and video processing,
computer vision, and video networking.

Dr. Lin is an IEEE Fellow. He has served as an Associate Editor of IEEE TIP, IEEE TCSVT, IEEE TMM, IEEE Multimedia, and Journal of Visual Communication and Image Representation. He was a Steering Committee member of IEEE TMM from 2014 to 2015. He is Distinguished
Lecturer of IEEE Circuits and Systems Society from 2018 to 2019. He also
serves as President of the Chinese Image Processing and Pattern Recognition Association, Taiwan, from 2019${\rm{ - }}$2020. He was Chair of the Multimedia
Systems and Applications Technical Committee of the IEEE Circuits and
Systems Society from 2013 to 2015. He served as Technical Program CoChair of IEEE ICME 2010, and will be the General Co-Chair of IEEE VCIP 2018 and Technical Program Co-Chair of IEEE ICIP 2019. His papers won
Best Paper Award of IEEE VCIP 2015, Top 10
MMSP 2013, and Young Investigator Award of VCIP 2005. He received the
Young Investigator Award presented by Ministry of Science and Technology,
Taiwan, in 2006.
\end{IEEEbiography}

\begin{IEEEbiography}[{\includegraphics[width=0.9in,height=1.2in]{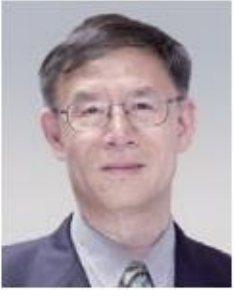}}]{David Zhang}
(Life  Fellow,  IEEE) graduated in computer science from Peking University in 1974. He received the M.Sc. and Ph.D. degrees in computer science from the Harbin Institute of Technology (HIT) in 1982 and 1985, respectively, and the Ph.D. degree in electrical and computer engineering from the University of Waterloo, Waterloo, ON, Canada. From 1986 to 1988, he was a Postdoctoral Fellow with Tsinghua University and then an Associate Professor with the Academia Sinica, Beijing. He has been a Chair Professor with The Hong Kong Polytechnic University, since 2005, where he is the Founding Director of the Biometrics Research Centre (UGC/CRC) supported by the Hong Kong SAR Government in 1998. He also serves as a Visiting Chair Professor with Tsinghua University and an Adjunct Professor with Peking University, Shanghai Jiao Tong University, HIT, and the University of Waterloo. He is the Founder and the Editor-in-Chief of the International Journal of Image and Graphics (IJIG), a Book Editor of the International Series on Biometrics (KISB), (Springer), an Organizer of the International Conference on Biometrics Authentication (ICBA), an ASSOCIATE EDITOR of more than ten international journals including the IEEE Transactions, and so on, and the author of more than ten books, over 300 international journal articles and 30 patents from USA, Japan, Hong Kong, and China. He is a Croucher Senior Research Fellow, a Distinguished Speaker of the IEEE Computer Society, a Fellow of IAPR and Fellow of Royal Society of Canada.
\end{IEEEbiography}
\end{document}